\PassOptionsToPackage{table, dvipsnames}{xcolor}
\documentclass[sigconf]{acmart}

\AtBeginDocument{%
  \providecommand\BibTeX{{%
    \normalfont B\kern-0.5em{\scshape i\kern-0.25em b}\kern-0.8em\TeX}}}

\copyrightyear{2024}
\acmYear{2024}
\acmDOI{XXXXXXX.XXXXXXX}

\newcommand{\ie}{\textit{i.e.,}~}
\usepackage{printlen}

\usepackage{array}
\usepackage{caption}
\usepackage{subcaption}
\usepackage{siunitx}
\uselengthunit{cm}
\usepackage{threeparttable}
\usepackage{fancyhdr}

\begin{document}

\title{Scalable and Effective Arithmetic Tree Generation \texorpdfstring{\\ for Adder and Multiplier Designs}{}}

\author{Yao Lai}
\email{ylai@cs.hku.hk}
\affiliation{%
  \institution{Department of Computer Science, \\The University of Hong Kong}
  \city{Hong Kong}
  \country{Hong Kong}
}

\author{Jinxin Liu}
\email{liujinxin@westlake.edu.cn}
\affiliation{%
  \institution{Machine Intelligence Lab, \\ Westlake University}
  \city{Hangzhou}
  \state{Zhejiang}
  \country{China}
}

\author{David Z. Pan}
\email{dpan@ece.utexas.edu}
\affiliation{%
  \institution{Department of Electrical \& Computer Engineering, \\ The University of Texas at Austin}
  \city{Austin}
  \state{Texas}
  \country{United States}
}

\author{Ping Luo}
\email{pluo@cs.hku.hk}
\affiliation{%
  \institution{Department of Computer Science, \\The University of Hong Kong}
  \city{Hong Kong}
  \country{Hong Kong}
}


\begin{abstract}
Across a wide range of hardware scenarios, the computational efficiency and physical size of the arithmetic units significantly influence the speed and footprint of the overall hardware system. Nevertheless, the effectiveness of prior arithmetic design techniques proves inadequate, as it does not sufficiently optimize speed and area, resulting in a reduced processing rate and larger module size.
To boost the arithmetic performance, in this work, we focus on the two most common and fundamental arithmetic modules: adders and multipliers. 
We cast the design tasks as single-player tree generation games, leveraging reinforcement learning techniques to optimize their arithmetic tree structures. 
Such a tree generation formulation allows us to efficiently navigate the vast search space and discover superior arithmetic designs that improve computational efficiency and hardware size within just a few hours. 
For adders, our approach discovers designs of 128-bit adders that achieve Pareto optimality in theoretical metrics.
Compared with the state-of-the-art PrefixRL, our method decreases computational delay and hardware size by up to 26\% and 30\%, respectively.
For multipliers, when compared to RL-MUL, our approach increases speed and reduces size by as much as 49\% and 45\%.
Moreover, the inherent flexibility and scalability of our method enable us to deploy our designs into cutting-edge technologies, as we show that they can be seamlessly integrated into 7nm technology.
We believe our work will offer valuable insights into hardware design, further accelerating speed and reducing size through the refined search space and our tree generation methodologies. 
See our introduction video at \href{https://bit.ly/ArithmeticTree}{\color{blue}{bit.ly/ArithmeticTree}}.
Codes are released at \href{https://github.com/laiyao1/ArithmeticTree}{\color{blue}{github.com/laiyao1/ArithmeticTree}}.
\end{abstract}



\maketitle

\begin{figure}[t]
  \centering
  \includegraphics[width=0.8\linewidth]{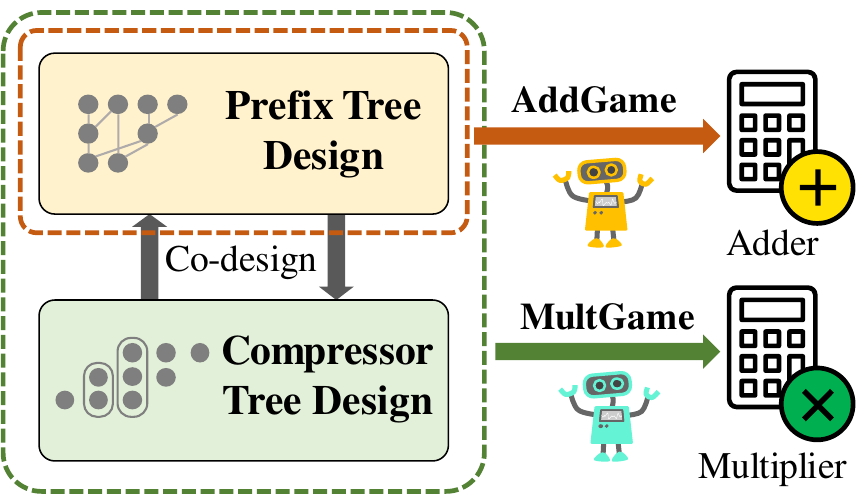}
  \caption{Our approach framework.
  Two agents respectively optimize prefix and compressor trees, modeling the tasks as AddGame for adders and MultGame for multipliers.
  }
  \label{teaser}
  \Description{}
\end{figure}

\section{Introduction}

\begin{figure*}[th]
  \centering
  \includegraphics[width=\linewidth]{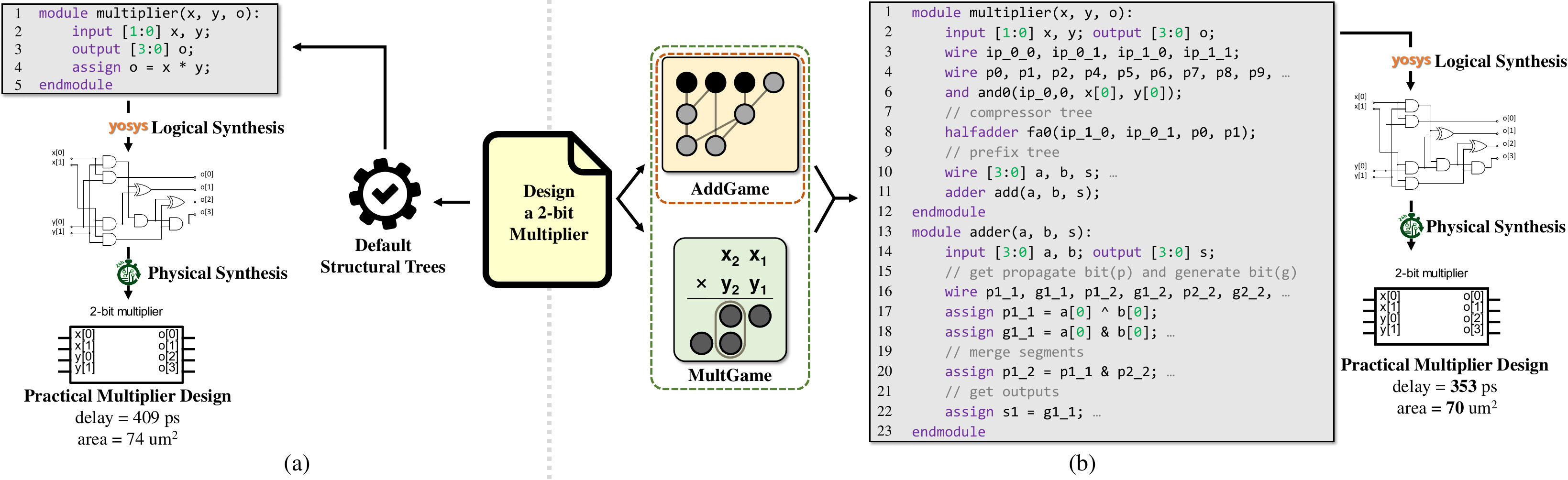}
  \caption{Comparison of design processes. (a) Default design process. 
  The synthesis tool automatically generates a default multiplier when directly using multiplication commands (x*y) in Verilog HDL code.
   (b) Our enhanced design process. 
   Our approach discovers an optimized multiplier structure and generates specialized Verilog HDL code for this improved structure, reducing delay and area after synthesis.
  }
  \label{design_process}
  \Description{}
\end{figure*}

Since the inception of computers, researchers have striven to boost computing speed and decrease hardware size. 
High computing speed is essential for a wide range of real-world applications, such as artificial intelligence~\cite{yao2020fully}, high-performance computing~\cite{haseeb2021high}, and high-frequency trading \cite{orus2019quantum}, particularly for the recent demanding applications of large language models like GPT~\cite{ouyang2022training}. 
Simultaneously, the push for smaller hardware has intensified with the rise of wearable devices and Internet of Things (IoT) technology~\cite{min2023autonomous}.

Over several decades, hardware specialists have continuously reduced the size of Complementary Metal-Oxide-Semiconductor (CMOS) technology \cite{bohr2017cmos} to enhance processor speeds and reduce chip footprints. 
However, as CMOS technology's scaling nears its fundamental physical limits \cite{taur2002cmos}, further miniaturization poses significant challenges.
Therefore, exploring innovative circuit design has emerged as a vital alternative to drive performance enhancement and area reduction.
Among the family of arithmetic modules for hardware architectures, 
adders and multipliers constitute two of the most essential modules, playing a critical role in various computational operations. 
For example, all convolution and fully connected layers of deep learning models are computed by basic addition and multiplication operations. 
Performance analysis of the ResNet-152 model \cite{he2016deep} reveals that the convolution operation, consisting solely of addition and multiplication, constitutes 98.4\% of the overall GPU execution time during model inference. 
Under Amdahl's Law \cite{rodgers1985improvements}, an enhancement of 30\% in addition and multiplication operation speeds could result in a 29\% improvement in inference speed.
Intriguingly, this improvement is comparable to the speedup typically seen with a generational upgrade in semiconductor process technology \cite{hiramoto2019five, salahuddin2018era}.
Therefore, designing more efficient and compact adders and multipliers is of paramount importance for the overall advancement of hardware design.

Since the 1960s, many arithmetic module design methods have been proposed.
These techniques generally fall into one of three main categories: human-based \cite{sklansky1960conditional, wallace1964suggestion}, optimization-based \cite{roy2013towards, roy2014towards, xiao2021gomil}, and learning-based \cite{roy2021prefixrl, dongsheng2023rl-mul, geng2021high}. 
However, these methods either demand significant hardware expertise or tend to get trapped in local optimal due to the vast design space for adders and multipliers modules. 
Specifically, for human-based methods, hardware experts have crafted a variety of arithmetic modules, such as the Sklansky adder \cite{sklansky1960conditional} and the Wallace multiplier \cite{wallace1964suggestion}. 
Nevertheless, as the number of input bits increases, designing new structures becomes increasingly challenging for humans. 
Optimization-based methods, like bottom-up enumerate search \cite{roy2013towards, roy2014towards} and integer linear programming \cite{xiao2021gomil}, can enhance the quality of arithmetic designs by exploring more irregular structures. 
Despite their potential, the extensive search space poses a challenge, necessitating manually defined assumptions to limit the search scope for feasible computation.
For example, Ma et al. \cite{ma2018cross} assumed the existence of semi-regular structures in adders, which may lead to locally optimal solutions. 
While learning-based approaches have emerged as a promising tool for automating hardware design in recent years \cite{roy2021prefixrl, dongsheng2023rl-mul, geng2021high}, navigating the vast design space to find the optimal solution for arithmetic hardware modules remains a formidable challenge. 
For example, the two primary components of an $N$-bit multiplier, the compressor tree and the prefix tree, have approximately $O(2^{N^2})$ and $O(2^{4N^2})$ design space \cite{roy2021prefixrl}, respectively. 
Consequently, the search space of a simple $16$-bit multiplier has already surpassed that of the Go game~($3^{361}$) \cite{silver2016mastering}. 
Meanwhile, such learning-based approaches also fail to consider the joint optimization of different components within arithmetic hardware \cite{roy2021prefixrl, dongsheng2023rl-mul}, thus easily leading to degenerated hardware with undesired performance bottleneck. 

To resolve the above limitations and boost the performance, 
we formulate the arithmetic adder and multiplier design problems as two single-player tree generation games, referred to as AddGame and MultGame, respectively, as shown in Fig.~\ref{teaser}. 
The key insight is that by reframing the design problems into interactive tree generation games, we harness the power of progressive optimization algorithms, allowing us to dynamically explore the intricate design space of arithmetic units. 
Starting from an initial prefix tree, 
the player in AddGame sequentially modifies cells in the prefix tree, in the same spirit as tactical movements in board games. 
Our MultGame contains two parts, specifically for designing the compressor tree and the prefix tree of multipliers. 
The compressor tree designing involves the player compressing all partial products with different compressors, similar to a match game, while the prefix tree designing follows the same rules as the AddGame. 
Unlike the default design process depicted in Fig. \ref{design_process}a, the tree structures discovered in games are converted into specific Verilog HDL codes~\cite{palnitkar2003verilog}, as illustrated in Fig. \ref{design_process}b.
We demonstrate that the delay and area of arithmetic modules can be largely decreased by substituting the default designs with our discovered tree structures.

In practical implementation, we utilize two customized reinforcement learning agents to optimize the prefix and compressor trees. 
For the prefix tree, appearing in both AddGame and MultGame, we utilize a Monte-Carlo Tree Search (MCTS) \cite{browne2012survey} agent to efficiently explore the large action space, while preserving previous exploration experience. 
For the compressor tree, exclusive to MultGame, we take a Proximal Policy Optimization (PPO) \cite{schulman2017proximal} agent due to its superior exploration efficiency. 
To capture the global design for multiplier designs, we also designed an optimization curriculum as depicted in Fig. \ref{teaser}, iteratively running MCTS and PPO agents to refine the prefix and compressor trees.

The main contributions of this paper are as follows: 
\begin{itemize}
  \item We model the arithmetic module design tasks as single-player tree generation games, \ie AddGame and MultGame, 
  which inherits the well-established RL capabilities for complex decision-making tasks (arithmetic tree optimization). 
  
  \item We propose the co-designed framework for
  prefix tree and compressor tree modules, 
  facilitating their collaboration to tackle more compound associations, and achieving the global optimal multiplier designs. 
  \item Our experiments reveal that our designed 128-bit Pareto-optimal adders outperform the latest theoretical designs. Also, our designed adders achieved up to 26\% and 30\% reductions in delay and area compared to PrefixRL \cite{roy2021prefixrl}, and multipliers offer 33\% and 45\% improvements over RL-MUL \cite{dongsheng2023rl-mul} in the same metrics. These designs are ready for direct integration into synthesis tools, offering significant industrial benefits, and are flexible and scalable enough to be seamlessly adopted into 7nm technology.
\end{itemize}

\begin{figure}[htbp]
  \centering
  \includegraphics[width=\linewidth]{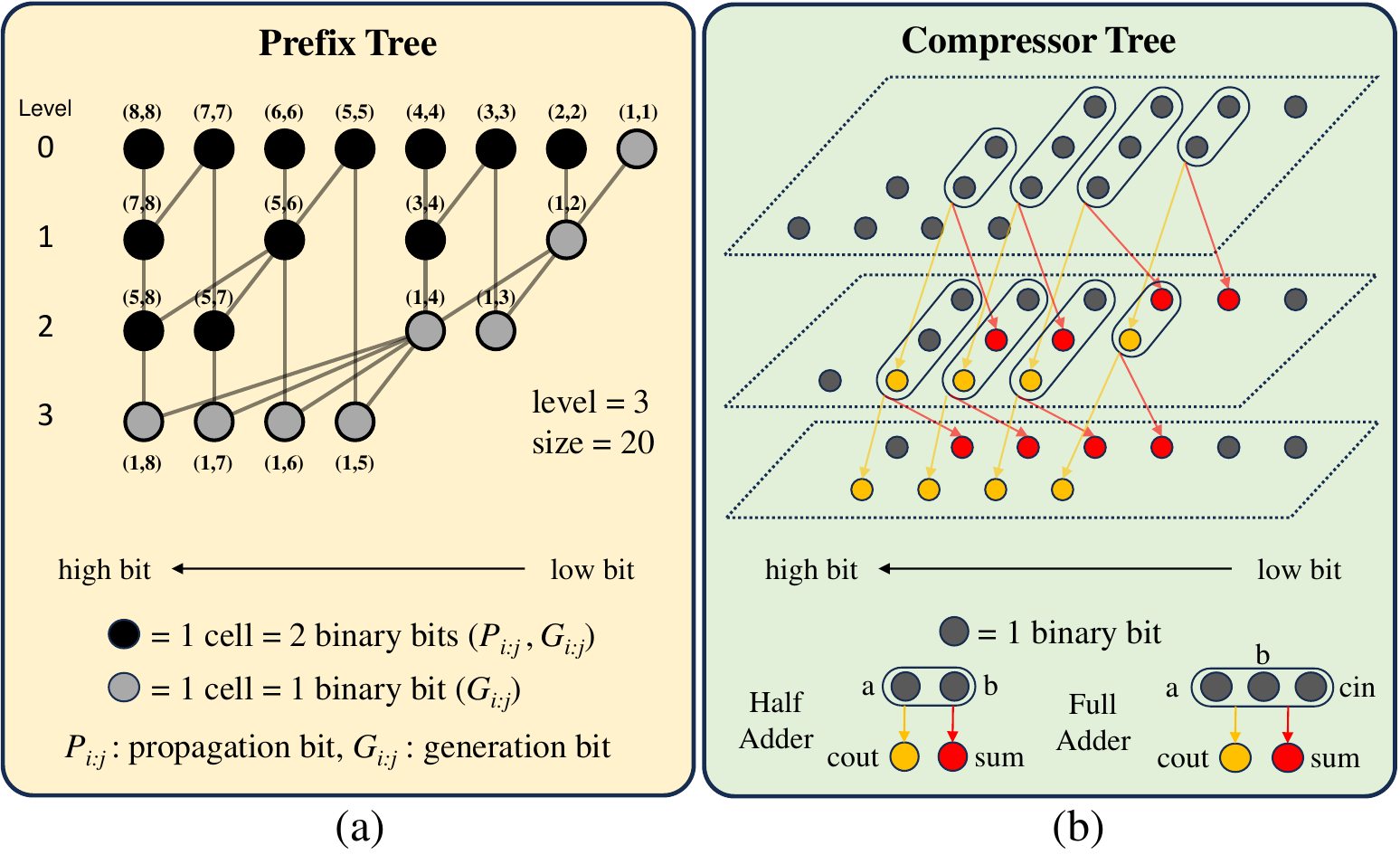}
  \caption{Arithmetic trees. (a) Example of a prefix tree. (b) Example of a compressor tree. Different tree structures lead to different qualities of adder and multiplier designs.
  }
  \Description{}
  \label{example}
\end{figure}

\section{Preliminaries}
\textbf{Adder Design.}
An $N$-bit adder can be constructed by cascading $N$ $1$-bit adders. 
However, this approach results in an $O(N)$ delay due to the sequential propagation of the carry signal from the lower bit to the higher bit. To address this issue, prefix adders have been proposed \cite{kogge1973parallel, ladner1980parallel}.
Prefix adders are designed based on the principles of addition, with a focus on reusing and parallelizing intermediate signal bits. 
These signal bits can be divided into two categories: propagation bits $p_i=a_i\oplus b_i$ and generation bits $g_i=a_i\cdot b_i$, where $a_i, b_i \in \{0, 1\}, i \in \{1, 2, \ldots, N\}$ represent the addends at the $i$-th bit, and ‘$\oplus$’ and ‘$\cdot$’ denote the logic XOR and AND operations, respectively \cite{roth2020fundamentals}.
These propagation and generation signals can be defined at both the individual bit level and across a range of bits. 
For an individual bit with index $i$, they are denoted by $P_{i:i}=p_i$ and $G_{i:i}=g_i$.
When considering a range of bits, this range is treated as an interval identified by a tuple $(i, j)$. 
Within each such interval, we have a single propagation signal $P_{i:j}=\prod_{k=i}^jp_k$ and a single generation signal $G_{i:j}=g_j+\sum_{k=i}^{j-1}P_{k:j}\cdot g_k$
, where `$+$' represents the logic OR operation.
Note that the computation of $P_{i:j}$ and $G_{i:j}$ is influenced solely by the input bits from position $i$ to $j$.
The ($N+1$) outputs of the adder can be calculated from the signal bits with the initial condition $G_{1:0}=0$ as follows:
\begin{align}
  c_{N+1}&=g_N + p_N \cdot G_{1:N}\\
  s_i&=p_i \oplus G_{1:i-1}
\end{align}
where $c_{N+1}$ is the carry-out bit and $s_i$ is the $i$-th sum bit.

The prefix adder design aims to optimize a hierarchical tree structure that generates all intervals $(1, i)$ from the initial intervals $(i, i)$, as shown in Fig. \ref{example}a. 
Signal bits for two adjacent intervals, $(i, k)$ and $(k+1, j)$, can be merged to form the larger interval $(i, j)$ by the computations: 
\begin{align}
    P_{i:j}&=P_{i:k}\cdot P_{k+1:j} \\
    G_{i:j}&=G_{i:k} \cdot P_{k+1:j} + G_{k+1:j}
\end{align}
This merging process generates a prefix tree.
In this tree, each cell represents one $(i, j)$ interval with two signal bits.
If an interval results from merging two others, its corresponding cell is the child node in the tree, and the merged intervals are its parent node.
For example, the $(5, 8)$ cell is the child node of the $(5, 6)$ and $(7, 8)$ cells because it derived from them.
A key advantage of this structure is that cells with no dependencies can be computed in parallel. 
Different tree structures can result in adders with varying delays and areas.
When evaluating the overall quality of the prefix adder, 
we can use the `level', corresponding to tree height, and the `size', denoting the number of cells, as substitutes for practical metrics delay and area, serving as the theoretical metrics. 

\textbf{Multiplier Design.}
An $N$-bit multiplier carries out the multiplication of two $N$-bit multiplicands, which can be regarded as the cumulative addition of $N$ addends, involving a total of $N^2$ bits.
Each addend represents a partial product with different powers of two weights, illustrated in Fig. \ref{ppo}b.
Multipliers can be easily achieved by cascading ($N-1$) $N$-bit adders or using a single $N$-bit adder ($N-1$) times.
However, both result in a large area or high delay.
To mitigate this, the $N^2$ bits in the partial products can be added simultaneously by $1$-bit adders, which can also be seen as a bit compression process because the number of bits gradually decreases.
The compression process halts when the number of bits for each binary digit is reduced to two or fewer before feeding into a downstream adder, as illustrated in Fig. \ref{example}b.
The process generates a compressor tree, describing a compression mechanism that merges $N^2$ bits into fewer bits by compressors such as half and full adders.
Introducing an additional carry-in input distinguishes a full adder from a half adder, as shown in Fig. \ref{example}b, which affects the latency and area. 
The difference is crucial when configuring the compressor tree in multipliers to optimize for delays and area requirements.
Upon completing the compression, the remaining bits are processed by a $2N$-bit prefix adder, designed to yield the globally optimal multiplier.

In summary, adder and multiplier design tasks can be interpreted as a tree-based structural generation process to optimize hardware metrics while maintaining functionality.

\section{Our Approach}
We solve the tree generation task for adder and multiplier designs by the reinforcement learning (RL) method.
The environments are modeled as single-player tree generation games, with AddGame dedicated to the design of adders and MultGame focused on creating multipliers, as illustrated in Fig. \ref{teaser}.
Taking into account the specific characteristics of the games, we propose an agent based on Monte Carlo Tree Search (MCTS) \cite{swiechowski2023monte} and another based on Proximal Policy Optimization (PPO) \cite{schulman2017proximal}.

\subsection{AddGame}
\begin{figure}[htbp]
  \centering
  \includegraphics[width=0.8\linewidth]{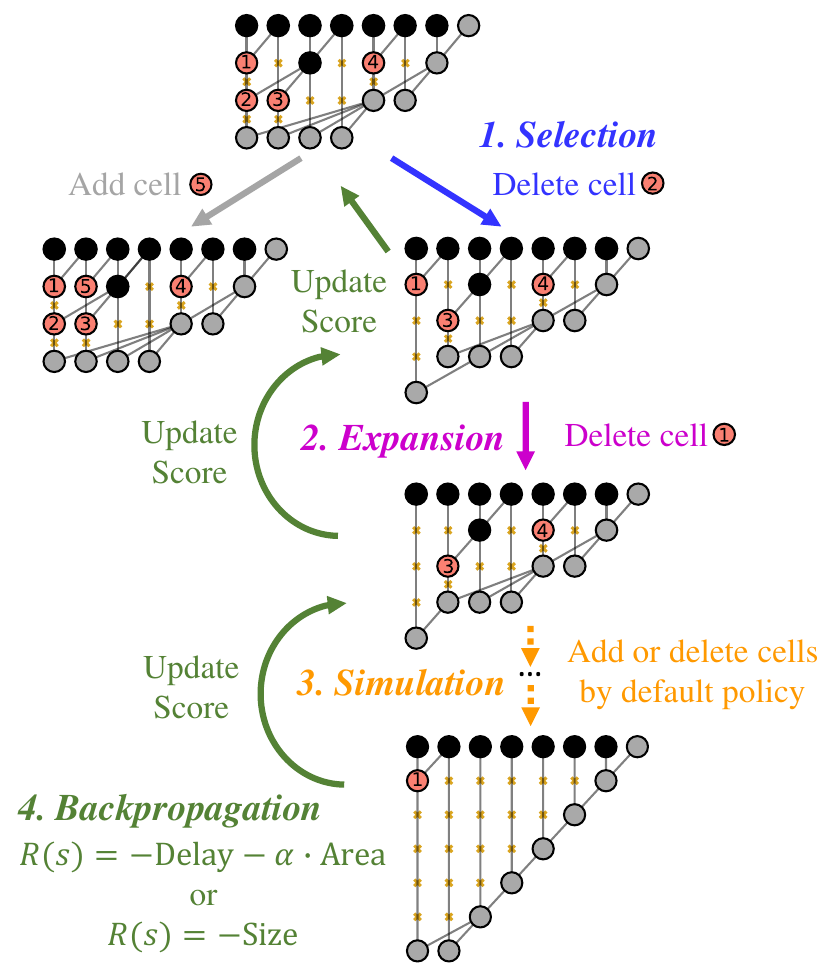}
  \caption{Method for designing prefix trees with MCTS. Four phases in the search process are executed iteratively, gradually building a search tree.}
  \Description{}
  \label{mcts}
\end{figure}

AddGame is modeled for designing prefix trees in adders and multipliers, as shown in Fig. \ref{mcts}.
In this game, the player modifies the structures of given initial prefix trees by basic actions to optimize the metrics of adders. 
The state of the game is denoted as $s$, corresponding to the current prefix tree.
In our evaluation, each state $s$ is assessed on two theoretical metrics: level and size, and two practical metrics: delay and area.
The player always chooses one action from two kinds of actions: (1) delete a cell $(i, j)$, (2) add a cell $(i,j)$, which $(i,j)$ is the cell index as shown in Fig. \ref{example}a. 
A cell $(i, j)$ $(i < j)$ can be deleted if the prefix tree does not have the cell $(i, k)$ subject to $k > j$ and $i > 1$, and all deletable cells are marked in red in Fig. \ref{mcts} and \ref{prefix_tree}.
A cell $(i, j)$ can be added if it does not exist in the prefix tree. 
All positions where can add cells are marked with `×'.
A legalization operation \cite{roy2021prefixrl} is always executed after one action to guarantee the feasibility of the prefix tree.
The game aims to maximize the performance score $R(s)$ of the adder $s$, which is determined by a weighted combination of delay and area (level and size when optimizing theoretical metrics).

Given the large action space, the agent for playing AddGame is based on an improved MCTS method, which has demonstrated its effectiveness in numerous game tasks \cite{silver2016mastering, vinyals2019grandmaster, fawzi2022discovering}.
Starting from the prefix trees in human-designed adders, 
the MCTS agent continuously cycles through four phases: \textit{selection}, \textit{expansion}, \textit{simulation}, and \textit{backpropagation}, and gradually builds a search tree in this process. Each node in the search tree represents one prefix tree.
\looseness=-1

In the \textit{selection} phase, the agent selects the child node with state $s$ that has the highest score $W(s)$, continuing until it encounters a node that has not been fully expanded.
The scores for evaluating nodes are computed by the Upper Confidence bounds applied to Trees (UCT) \cite{kocsis2006bandit}, keeping the balance between exploration and exploitation.
In the search tree, each node with the state $s$
stores a visit count $N(s)$ and an action value $V(s)$.
The visit count $N(s)$ records the number of visits to the node $s$. 
The action value $V(s)$ is the weighted sum of the best performance score $\max{R}$ and average performance score $\overline{R}$ of all its descendant nodes, which can be formalized as:

\begin{align}
V(s) = (1-\beta)\underbrace{\sum_{s'\in D(s)}R(s')/| D(s)|}_{\text{avg performance score}} \  + \ \beta\underbrace{\max_{s'\in D(s)}R(s')}_{\text{best performance score}}
\label{action_value}
\end{align}
where $D(s)$ represents all descendant nodes of the node $s$ (including $s$ itself), \ie all generated adders by a sequence of actions from adder $s$. $|\cdot|$ gives the number of nodes. 
$R(s')$ indicates the performance score of the adder of the state $s'$, which is defined as $-\text{Delay}-\alpha\text{Area}$ or $-\text{Size}$.
$\alpha$ and $\beta$ are sum weights.

We define the node score $W(s)$ with the state $s$ as follows:

\begin{align}
W(s) = \sqrt{\frac{\text{ln}{N(P(s))}}{N(s)}} + cV(s)
\label{value}
\end{align}
where $P(s)$ is the parent node of $s$, $N(\cdot)$ is visit count function,
and $c$ is an adjustable parameter.

In the \textit{expansion} phase, a random action is chosen from the unexplored actions available at the node identified in the selection phase and executed. It expands the search tree by adding a new node corresponding to the result after that action. 

In the \textit{simulation} phase, a sequence of actions is taken until the performance scores of adders can no longer be improved (in theoretical metrics optimization) or the simulation exceeds the maximum steps (in practical metrics optimization).

In the \textit{backpropagation} phase, the last state $s$ reached in the simulation phase is evaluated to get a performance score $R(s)$, which is then backpropagated to update the scores of all preceding nodes in the search tree according to Eq. \ref{action_value} and \ref{value}.

\textbf{Pruning.}
To enhance efficiency, we implement pruning techniques to avoid the exploration of unnecessary sub-trees. 
When optimizing theoretical metrics, we restrict modifications to `delete cell' actions, as adding cells does not improve the design outcome.
Furthermore, we impose an upper limit on the depth of the prefix trees to prevent the creation of structures with excessively high complexity. 
This upper limit, denoted as $L$, is set for each MCTS search and is gradually relaxed for each search iteration.

\textbf{Two-level Retrieval.} 
We adopt a two-level retrieval strategy to balance synthesis accuracy and computational efficiency, dividing the search into two stages because the full synthesis flow is highly accurate but time-consuming.
A faster yet marginally less simulating accurate synthesis flow is employed in the first stage, eliminating the time-intensive step.
Only the top $K$ adders identified in the first stage undergo full synthesis in the second stage.

\begin{figure*}[th]
  \centering
  \includegraphics[width=\linewidth]{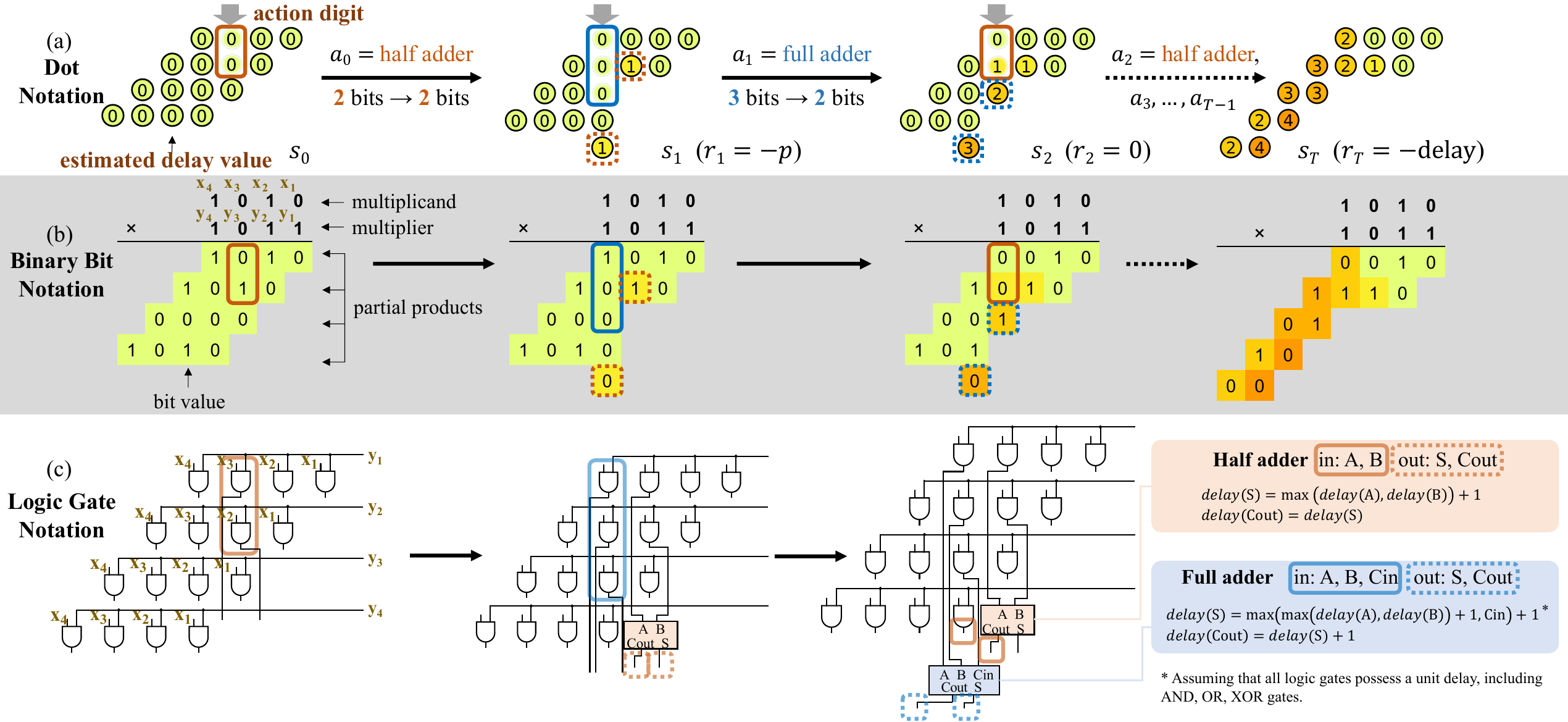}
  \caption{Method for designing compressor trees with PPO. 
  Three representations are illustrated.
  (a) Dot notation.
  Each dot represents an output bit, with the number inside indicating the estimated delay for selecting adder input bits. 
  The agent's actions involve adding full or half adders to compress the bits until each binary digit contains no more than two bits. 
  The final reward, $r_T$, is defined as the inverse of the delay, encouraging designs with lower delays.
(b) Binary bit notation. 0/1 are values of bits for the example multiplication.
(c) Logic gate notation. The actual logic gate circuit design for each state.
}
  \label{ppo}
  \Description{}
\end{figure*}

\subsection{MultGame}
MultGame consists of two parts for jointly designing compressor and prefix trees in multipliers as shown in Fig. \ref{teaser}.
The part dedicated to the design of the prefix tree is identical to that in AddGame, whereas the section for designing the compressor tree involves continually merging bits in partial products through compression actions, as depicted in Fig. \ref{ppo}, similar to some match games like `2048' \cite{dedieu2017deep} to merge items in a specific way to get high scores.

The compressor tree is built from scratch instead of starting from existing solutions for more design flexibility.
The game state $s_t$ at step $t$ is represented by a vector representing the current compressor tree status.
The player chooses one of two actions: (1) using a half adder or (2) using a full adder to compress bits at the action digit, which is defined as the lowest digit containing more than two bits, as indicated in Fig. \ref{ppo}a.
Half and full adders compress two or three bits in the $k$-th digit and generate a carry-out bit in the ($k+1$)-th digit and a sum bit in the $k$-th digit.
Rough delays for all bits are estimated, assuming a one-unit delay for all basic logic gates, as shown in the dots of Fig. \ref{ppo}a.
To minimize the increase in total delay, the bits with the lowest estimated delays are selected as inputs for the adders.
The game terminates at step $T$ when all digits have two or fewer bits.
A reward $r_T$ is computed through the synthesis tools as the negative of the delay, denoted $r_T = -\text{delay}$.
Moreover, a penalty term $-p$ is also applied to $r_t$ if the action $a_{t-1}$ uses a half adder, where $1\leq t \leq T$.
This penalty reflects the fact that a full adder accepts three input bits (two addend bits and a carry-in bit) and produces two output bits (a sum bit and a carry-out bit), effectively reducing the bit count. 
In contrast, a half adder only processes two addend bits and outputs two bits, thus not contributing to a reduction in bit count. 
A half adder's lack of bit count reduction can lead to more adder modules, increasing the overall area.

We train an RL agent using the PPO method with policy and value networks. 
Both networks are built by multi-layer perceptions (MLPs) \cite{murtagh1991multilayer} with three layers. 
The inputs comprise pre-defined features such as action digit, max delay, number of half adders, eligible action type, and the estimated delays of bits.
The policy and value networks contain $(64,16,2)$ and $(64,8,1)$ neurons in each layer. 
The last layer of the policy network is connected to a Softmax activation function \cite{bishop2006pattern} to make it a binary classification task for choosing the action.
When training, the objective function can be defined as follows for maximizing the game's cumulative reward:

\begin{align}
J(\theta) = \mathbb{E}_{\tau\sim \pi_{\theta}}\big[G_T\big] \
= \mathbb{E}_{\tau\sim \pi_{\theta}}\big[\sum_{i=0}^{T}\gamma^i r_i\big]
\label{loss_func}
\end{align}
where $\tau = (s_0, a_0, s_1, r_1, a_1, ... , a_{T-1}, s_T, r_T)$ is a trajectory from the game episode, 
and $\pi_{\theta}$ denotes the policy parameterized by $\theta$.
$G_T$ refers to the cumulative discounted reward from step 0 to step $T$.
The discount factor, $\gamma$, is for emphasizing immediate rewards rather than future rewards.
When implementing the PPO, the objective function for optimizing the policy network
can be formalized~as:

\begin{align}
L(\theta) = \hat{{\mathbb{E}}}_t\big[\min\big(r_t(\theta)\hat{A}_t,\ \text{clip}(r_t(\theta), 1-\epsilon, 1+\epsilon)\hat{A}_t \big) \big]
\label{loss}
\end{align}
where $\hat{\mathbb{E}}_t[\cdot]$ indicates the empirical average over a finite batch of samples, and $r_t(\theta)$ denotes the probability ratio $\frac{\pi_\theta(a_t|s_t)}{\pi_{\theta_{\mathrm{old}}}(a_t|s_t)}$. Here, $\theta_{\mathrm{old}}$ is the policy network parameters before the update.
$\hat{A}_t = G_t - \hat{V}_t$ is an estimation of the advantage function at step $t$, and $\hat{V}_t$ is the value estimated by the value network. $\text{clip}(\cdot, 1-\epsilon, 1+ \epsilon)$ is the function restricting results to the interval $[1-\epsilon, 1+ \epsilon]$.

Simultaneously, the value network with parameters $\phi$ is updated by optimizing the following objective function:

\begin{align}
L(\phi) = \hat{{\mathbb{E}}}_t\big[\text{smooth\_L1}(G_t, \hat{V}_t) \big]
\end{align}
where $\text{smooth\_L1}(\cdot)$ is the smooth L1 loss function \cite{girshick2015fast}.

\textbf{Synthesis Acceleration.}
In RL-MUL~\cite{dongsheng2023rl-mul}, running synthesis tools proved to be a bottleneck, especially for scaling to multipliers with higher bit-widths. To address this, our enhancements to the synthesis flow yield a 10x speedup in reward computation without sacrificing accuracy. These modifications facilitate the design of multipliers up to $64$-bit, expanding from the previous $16$-bit limit. Enhancements include activating the fast mode in the logical synthesis script and adopting direct code template-based generation of Verilog HDL code from our search results, moving away from the EasyMAC \cite{zhang2022easymac} tool.

\textbf{Co-design Framework.}
As shown in Fig. \ref{teaser}, we have developed a joint design approach that simultaneously optimizes the two primary components of the multiplier: the prefix and compressor trees. Our method involves an iterative process, where each round consists of optimizing the compressor tree with a fixed prefix tree, followed by searching for an ideal prefix tree that aligns with the optimized compressor. This alternating optimization continues until the computational iterations conclude.

More details can be seen in the Appendix \ref{method_detail}.

\begin{figure*}[th]
\centering
  \includegraphics[width=\linewidth]{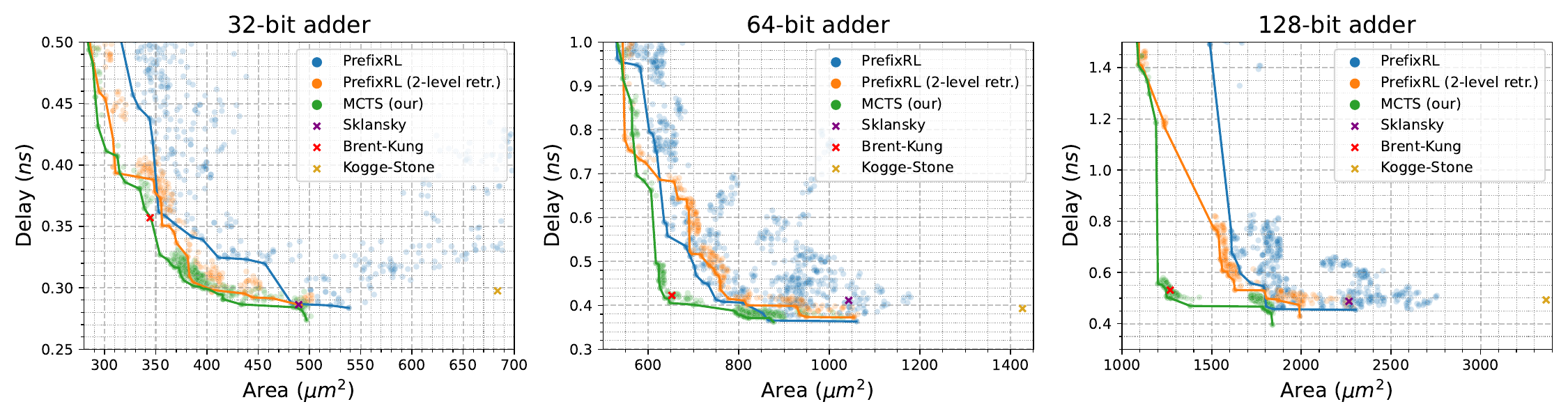}
  \caption{Comparison of adders in delay and area. 
Each point represents one discovered adder, and line segments connect Pareto-optimal adders for each method.
`PrefixRL (2-level retr.)' is the raw PrefixRL method improved by our two-level retrieval strategy.
Sklansky, Brent-Kung, and Kogge-Stone refer to human-designed adders.
The MCTS method can significantly improve the delay and area, particularly for high-bit adders.
Furthermore, it can discover adders with minimal delays. 
Our two-level retrieval strategy can effectively find superior designs by expanding the sampling size.}
  \label{adder_all}
  \Description{}
\end{figure*}

\section{Experiments}

\subsection{Experimental Settings}

We use the logic synthesis tool Yosys $0.27$ version \cite{wolf2016yosys}, and the physical synthesis tool OpenROAD $2.0$ version \cite{ajayi2019openroad} to test the delay and area of the designed adders and multipliers with technology process libraries, Nangate45 \cite{nangate2008nangate} and ASAP7 \cite{clark2016asap7}. 
Both synthesis tools are open-sourced for the reproduction of results.
All experiments are run on one GeForce RTX 3090 GPU and one AMD Ryzen 9 5950X 16-core CPU.
We give our detailed design synthesis scripts and hyperparameter settings in Appendix  \ref{synthesis_script} and \ref{hyperparameter}.

\subsection{Adder Design}

\begin{figure}[!ht]
  \centering
  \includegraphics[width=\linewidth]{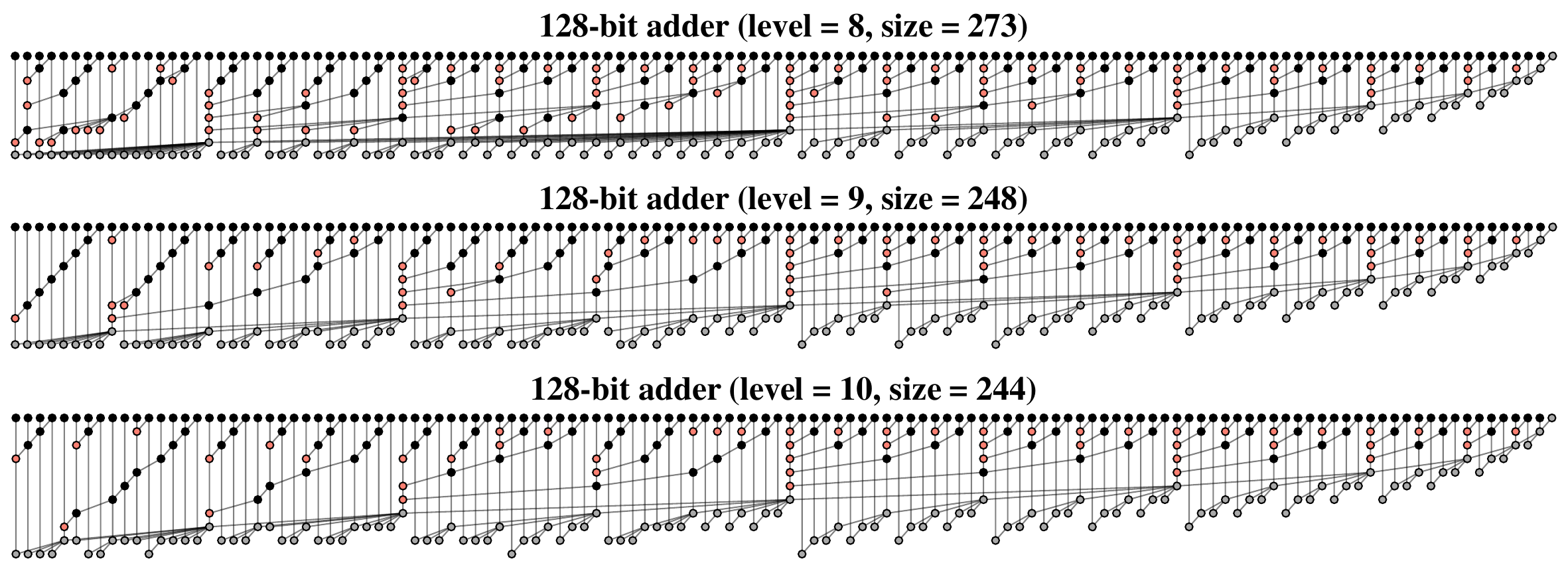}
  \caption{Some first discovered prefix trees for 128-bit adders with the smallest sizes.
  }
  \Description{}
  \label{prefix_tree}
\end{figure}

\textbf{Theoretical Evaluation.} 
As illustrated in Fig. \ref{example}a, prefix tree structures determine the theoretical metrics, including level and size, which strongly correlate with delay and area, and these metrics are independent of specific technologies.
From our observations, optimizing the level presents more challenges than size.
Therefore, when optimizing theoretical metrics, we make the search objective is to find the optimal size for each given level upper bound $L$ in MCTS.
We initiate our search with the Sklansky adder \cite{sklansky1960conditional}, which possesses a theoretical minimum level of $log_{2}N$.
We initially set $L$ to this minimum value and increase it incrementally by one in subsequent searches. 
For each new iteration, the initial state is the prefix tree with the minimal size identified in the preceding search.
We limited the number of steps to $4 \times 10^5$ for each search iteration. 
For comparative purposes, we replicated the PrefixRL~\cite{roy2021prefixrl} by optimized hyperparameters and an identical number of steps. 
The results for other methods were obtained directly from the respective original publications.
The results in Table \ref{theory} indicate that our method surpasses the state-of-the-art designs documented ten years ago in \cite{roy2013towards}.
We present some adder structures we discovered in Fig. \ref{prefix_tree}.
Despite the exponentially growing search space, our MCTS method can enhance $128$-bit adders, surpassing the designs from optimization-based methods.
Notably, guided by Snir's theoretical lower bound for size at a given level \cite{snir1986depth}, we were the first to discover a $128$-bit adder with $10$ levels and a size of $244$, which is verified as theoretically optimal within this level.
\looseness=-1

\begin{table}[!ht]
    \centering
    \caption{
    Comparisons of adders in size and area. The smaller the size and level, the better.
    }
    \label{theory}
    \small
    \begin{tabular}{>{\centering\arraybackslash}p{0.7cm}>{\centering\arraybackslash}p{0.7cm}|>{\centering\arraybackslash}p{1cm}|>{\centering\arraybackslash}p{0.9cm}>{\centering\arraybackslash}p{0.9cm}>{\centering\arraybackslash}p{0.9cm}>{\centering\arraybackslash}p{0.9cm}}
     \toprule
        Input Bit & Level & Theory Size Bound \cite{snir1986depth}  & Sklansky Size \cite{sklansky1960conditional} & Area Heuristic \cite{matsunaga2007area} & Best Known Size \cite{roy2013towards} & Our Size \\ \midrule
        64 & 6 & 120 & 192 & 169 & \textbf{167} & \textbf{167}  \\ 
        64 & 7 & 119 & - & 138 & \textbf{126} & \textbf{126}  \\ 
        64 & 8 & 118 & - & 120 & \textbf{118} & \textbf{118}  \\ 
        64 & 9 & 117 & - & \textbf{117} & \textbf{117} & \textbf{117}  \\ 
        64 & 10 & 116 & - & \textbf{116} & \textbf{116} & \textbf{116}  \\ \midrule
        128 & 7 & 247 & 448  & 375 & \textbf{364} & \textbf{364}  \\ 
        128 & 8 & 246 & -  & 304 & 276 & \textbf{273}  \\ 
        128 & 9 & 245 & -  & 284 & 250 & \textbf{248}  \\ 
        128 & 10 & 244 & -  & 257 & 245 & \textbf{244}  \\ \bottomrule
    \end{tabular}
\end{table}

\textbf{Practical Evaluation.} We compute practical metrics, including the delay and area of hardware modules, through synthesis tools for evaluation.
We run $1000$ full syntheses for adders in each method to ensure a fair comparison. 
Our MCTS agent begins each search from one of three different types of adders: the Sklansky adder \cite{sklansky1960conditional}, Brent-Kung adder \cite{brent1982regular}, and ripple-carry adder \cite{behrooz2000computer}, chosen uniformly to leverage their respective strengths.
A two-level retrieval strategy is implemented by dividing the search into two stages: 
(1) run $5000$ fast syntheses. 
(2) run $500$ full syntheses with the top $500$ adders selected from the first stage. 
Efficiency tests show that the computational load of one full flow equals that of ten fast flows. Thus, the proposed strategy achieves the same computational volume by combining the load from $5000$ fast flows, at $0.1$ units each, with $500$ units from full flows, equating to the load of $1000$ full flows.
The state-of-the-art method PrefixRL \cite{roy2021prefixrl} is implemented with its optimal settings.
In our results in Fig. \ref{adder_all}, each prefix adder is represented by a 2D point based on its delay and area.
It shows the significant improvement achieved when our two-level retrieval strategy is used in the PrefixRL method, due to efficiency improvements that facilitate the exploration of an expanded corpus of samples.
Moreover, when employing the MCTS method, one can find more superior adders, because this method effectively navigates problems with vast state spaces, utilizing information stored during the search process.
Overall, our approach can reduce the delay or area of adders by up to 26\% and 30\% respectively compared with PrefixRL, while maintaining the computational amount.

\begin{figure}[!ht]
  \centering
  \includegraphics[width=\linewidth]{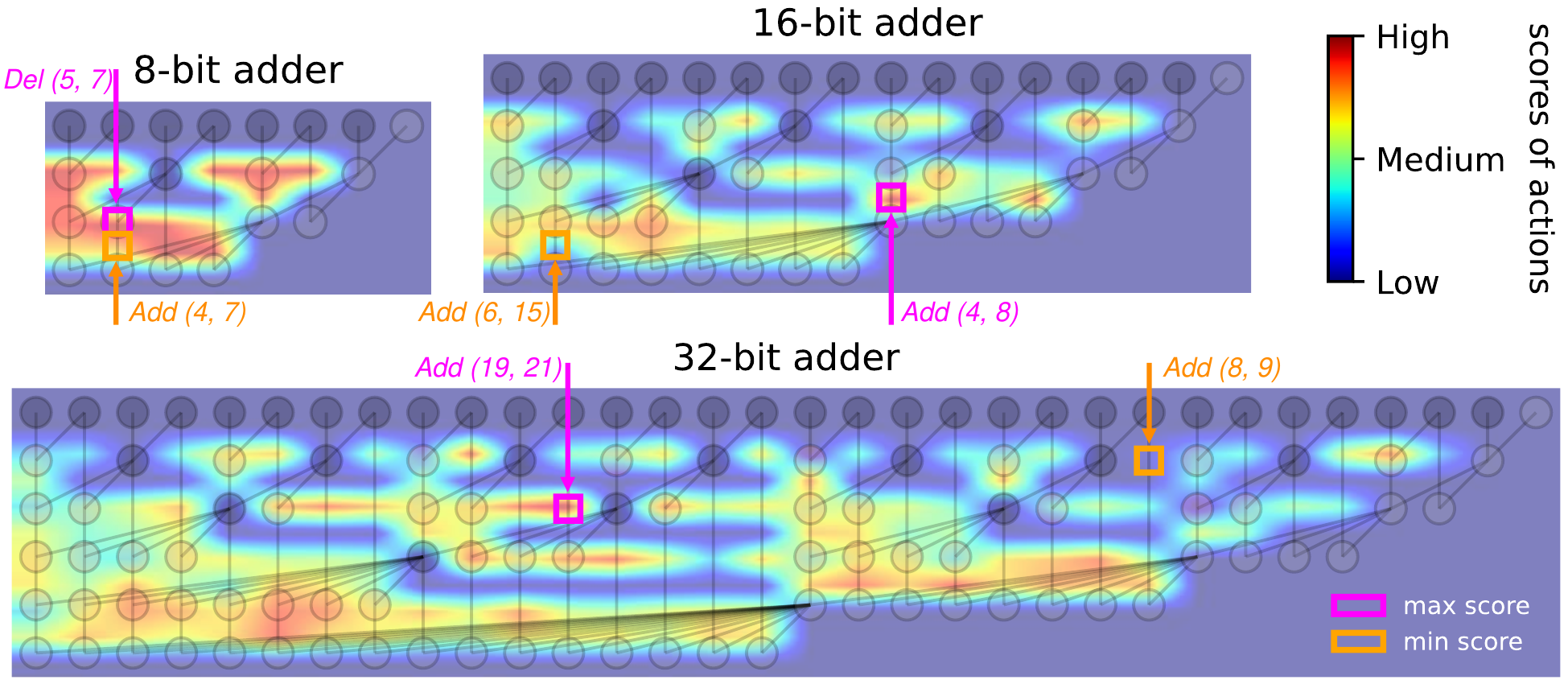}
  \caption{Heatmap for first action scores.
The actions with the highest and lowest scores are marked.
  }
  \Description{}
  \label{heatmap}
\end{figure}

\textbf{Visualization.}
The scores of all first actions after $400$ search steps when optimizing the theoretical metrics are visualized as heatmaps in Fig. \ref{heatmap}. 
These scores computed by Eq. \ref{value} provide a balanced assessment for exploration and exploitation. 
In the selection phase of MCTS, the action with the highest score is chosen.
Scores are visualized by colors at the corresponding action positions. 
Warmer colors on the heatmap denote actions with higher scores. 
For example, the first action for the $8$-bit adder is to delete the $(5, 7)$ cell with the highest score, primarily because this reduces the size of the adder. 
On the contrary, the action with the lowest score is to add the $(4, 7)$ cell because it concurrently augments both size and level.

\begin{figure*}[th]
\centering
  \includegraphics[width=\linewidth]{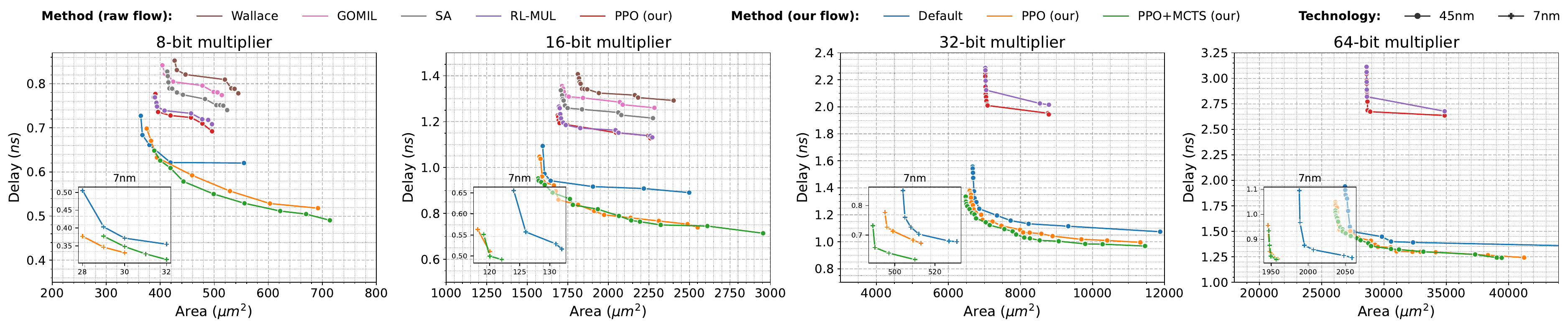}
  \caption{Comparison of multipliers in delay and area. 
The designs were tested in 45nm and 7nm processes.
Each segmented line represents the performance of one multiplier under different timing constraints.
Methods categorized under `our flow' utilize our improved synthesis flow.
The `Default' multipliers are those generated by the synthesis tool by default.
`PPO+MCTS (our)' refers to our co-design method, while `PPO (our)' optimizes the compressor tree only.
We apply 45nm designs to the 7nm library without modifications, showcasing the high transferability.
}
  \Description{}
  \label{multiplier}
\end{figure*}

\begin{table*}[!ht]
    \caption{Numerical comparison of multipliers in delay and area with two objectives}
    \small
    \label{numerical}
    \centering
    \begin{tabular}{l|l|cc|cc|cc|cc}
    \toprule
         ~ & Num of bits & \multicolumn{2}{c|}{8-bit} & \multicolumn{2}{c|}{16-bit} & \multicolumn{2}{c|}{32-bit} & \multicolumn{2}{c}{64-bit}   \\ \midrule
        Objective & Method & area (\SI{}{\micro\metre\squared}) & delay (ns)& area (\SI{}{\micro\metre\squared}) & delay (ns) & area (\SI{}{\micro\metre\squared}) & delay (ns) & area (\SI{}{\micro\metre\squared}) & delay (ns) \\ \midrule
        ~ & RL-MUL & 496 & 0.7089 & 2271 & 1.1330 & 8767 & 2.0150 & 34810 & 2.6771  \\ 
        ~ & PPO w/ raw flow & 496 & 0.6921 & 2259 & 1.1277 & 8788 & 1.9437 & 34810 & 2.6355  \\ 
        Min Delay & Default & 555 & 0.6203 & 2499 & 0.8908 & 10637 & 1.0745 & 42128 & 1.3498  \\ 
        ~ & PPO & 692 & 0.5180 & 2551 & 0.7392 & 11329 & 0.9960 & 41237 & 1.2424  \\ 
         ~ & \cellcolor{gray!25}PPO+MCTS & \cellcolor{gray!25}714 & \cellcolor{gray!25}\textbf{0.4905} & \cellcolor{gray!25}2955 & \cellcolor{gray!25}\textbf{0.7138} & \cellcolor{gray!25}11460 & \cellcolor{gray!25}\textbf{0.9685} & \cellcolor{gray!25}39436 & \cellcolor{gray!25}\textbf{1.2401}  \\ \midrule
        ~ & RL-MUL & 388 & 0.7691 & 1695 & 1.2668 & 7033 & 2.1932 & 28616 & 2.8891  \\ 
        ~ & PPO w/ raw flow & 388 & 0.7618 & 1687 & 1.2268 & 7036 & 2.0945 & 28609 & 2.8928  \\ 
        Trade-off & Default & \textbf{367} & 0.6837 & 1590 & 0.9997 & 6685 & 1.4170 & 26871 & 1.9403  \\ 
        ~ & PPO & 377 & 0.6558 & 1568 & 1.0135 & 6581 & 1.3856 & 26088 & 1.7941  \\ 
        ~ & \cellcolor{gray!25}PPO+MCTS & \cellcolor{gray!25}384 & \cellcolor{gray!25}\textbf{0.6420} & \cellcolor{gray!25}\textbf{1566} & \cellcolor{gray!25}\textbf{0.9487} & \cellcolor{gray!25}\textbf{6469} & \cellcolor{gray!25}\textbf{1.3262} & \cellcolor{gray!25}\textbf{26087} & \cellcolor{gray!25}\textbf{1.7038} \\ \bottomrule
    \end{tabular}
\end{table*}

\subsection{Multiplier Design}

\textbf{Practical Evaluation.}
Given the lack of a commonly adopted theoretical metric for multipliers, we introduce the delay and area as practical metrics for comparison.
The multiplier design is based on a co-design framework alternating between optimizing the compressor and prefix trees.
This iterative search is conducted over three rounds, optimizing each component three times. 
In each round, we perform 900 search steps on the compressor tree and 100 steps on the prefix tree, resulting in a total of 3000 steps across all three rounds, which aligns with the step count of other methods.
As shown in Fig. \ref{multiplier}, we compared the effectiveness of our method with several baselines, including the human-designed Wallace multiplier \cite{wallace1964suggestion}, optimization-based methods including GOMIL \cite{xiao2021gomil} and SA \cite{van1987simulated}, the default multiplier given in the synthesis tool, and the learning-based method RL-MUL \cite{dongsheng2023rl-mul}.
In our evaluation, we assessed the performance of multipliers by adjusting the expected delay parameter in the synthesis process. 
Subsequently, the resulting areas of each multiplier at different delays are depicted as a segmented line.
Consistent with the RL-MUL \cite{dongsheng2023rl-mul} assessment approach, each method selects an optimal multiplier for comparison.
Results for Wallace~\cite{wallace1964suggestion}, GOMIL~\cite{xiao2021gomil}, SA~\cite{van1987simulated}, and RL-MUL~\cite{dongsheng2023rl-mul} in 8/16 bits are referenced from the RL-MUL work.
RL-MUL method is reproduced and tested in 32/64 bits.
According to the results, our co-design method, PPO+MCTS, outperforms the baselines and the default multipliers of the synthesis tool. 
This is because of the co-design framework, the restructured MultGame, and the improved synthesis flow.
It can achieve second-best results even when only optimizing the compressor tree. 
Compared with the state-of-the-art RL-MUL method, our method can reduce the delay by up to 33\% (equivalent to a speed increase of 49\%) and the area by 45\%. 
Furthermore, our method can reduce the delay of the default multipliers used in the Yosys synthesis tool \cite{wolf2016yosys} by up to 16\% (equivalent to a speed increase of 19\%) and the area by 35\%.

We also report the values of delay and area with 45nm technology in Table \ref{numerical}. Our method consistently achieves minimal delays for the delay minimization objective.
When optimizing for a trade-off between delay and area (objective function: $\text{delay} + 0.001\text{area}$), our approach achieves optimal or comparable results.

\textbf{Technology Transfer.}
We have verified that the multipliers discovered using 45nm technology are compatible with the 7nm technology library \cite{clark2016asap7}. 
These multipliers, when applied directly to this more advanced 7nm technology, demonstrate optimal performance, as shown in Fig. \ref{multiplier}, without necessitating any modifications. 
This suggests that our method will likely remain effective as semiconductor process technology advances.

\subsection{Efficiency}
Due to the time-consuming nature of full synthesis flow, we developed a fast synthesis flow that is over 10 times faster while maintaining high simulation accuracy for the adder design.
The efficiency is shown in Fig. \ref{efficiency}a, and the accuracy is verified in Appendix Table \ref{simplified_synthesis}.
Additionally, we optimized the logic synthesis and HDL code generation processes in the synthesis flow for multiplier design. According to Fig. \ref{efficiency}b, our improved fast flow can accelerate the process by up to 20 times.

\begin{figure}[ht]
  \centering
  \includegraphics[width=\linewidth]{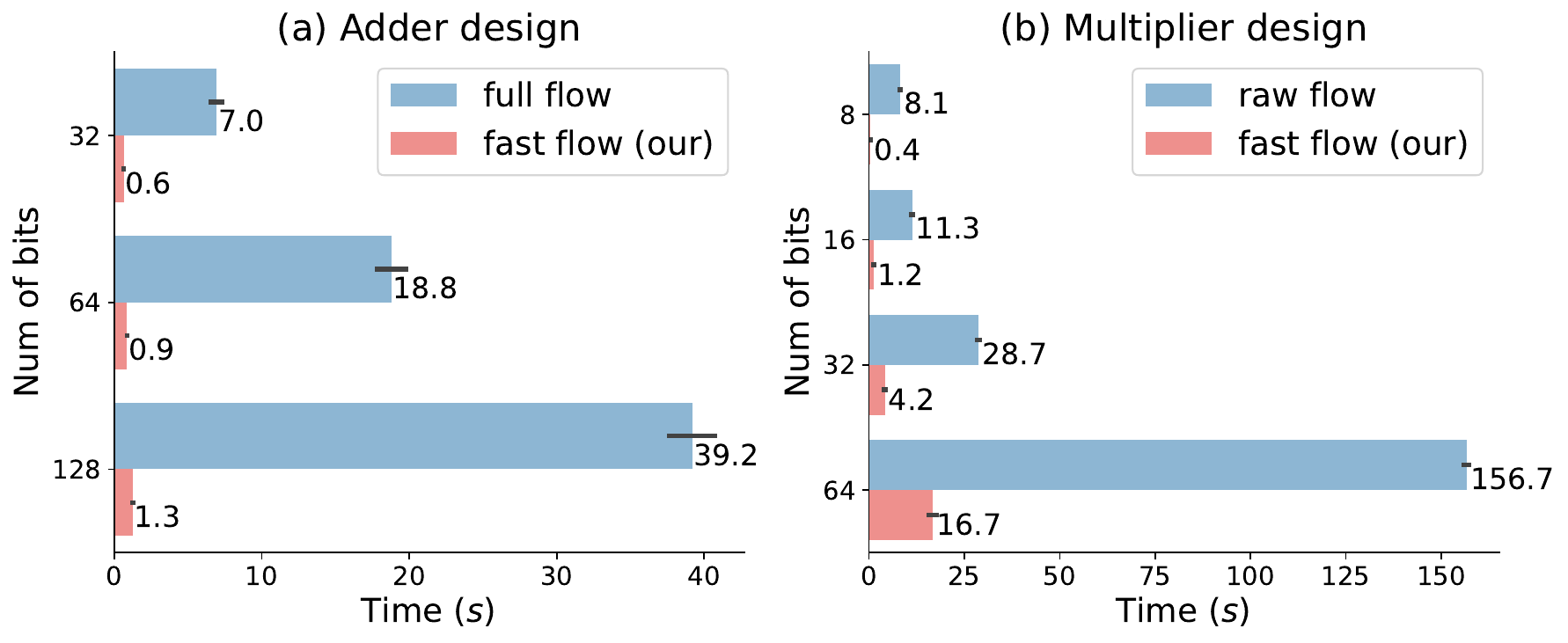}
  \caption{Time consumption for one design flow
  }
  \Description{}
  \label{efficiency}
\end{figure}

\section{Conclusion}
Designing adder and multiplier modules is a fundamental and crucial task in computer science. 
We first model the task as a tree generation process, which we conceptualize as a sequential decision-making game, and then propose a reinforcement learning method to solve it,
facilitating a scalable and efficient search for globally optimal designs. 
Through extensive experiments, we demonstrate that our approach achieves state-of-the-art performance for adders and multipliers regarding delay and area within the same computational resources.
Moreover, our method has demonstrated transferability as our discovered designs can be applied to the more advanced 7nm technology process. 

We believe that our work will be of interest to a broad academic and industry community. Given our approach's simplicity and generality, our work can serve as a foundation for future works on understanding and improving hardware designs, including more complex hardware modules. The improvement in basic arithmetic modules also carries the potential to significantly improve the current hardware's performance and size, which is beneficial for many fields in human society, such as artificial intelligence and high-performance computing.

\newpage

\bibliography{sample-base}

\begin{thebibliography}{10}

\bibitem{yao2020fully}
Peng Yao, Huaqiang Wu, Bin Gao, Jianshi Tang, Qingtian Zhang, Wenqiang Zhang, J~Joshua Yang, and He~Qian.
\newblock Fully hardware-implemented memristor convolutional neural network.
\newblock {\em Nature}, 577(7792):641--646, 2020.

\bibitem{haseeb2021high}
Muhammad Haseeb and Fahad Saeed.
\newblock High performance computing framework for tera-scale database search of mass spectrometry data.
\newblock {\em Nature computational science}, 1(8):550--561, 2021.

\bibitem{orus2019quantum}
Rom{\'a}n Or{\'u}s, Samuel Mugel, and Enrique Lizaso.
\newblock Quantum computing for finance: Overview and prospects.
\newblock {\em Reviews in Physics}, 4:100028, 2019.

\bibitem{ouyang2022training}
Long Ouyang, Jeffrey Wu, Xu~Jiang, Diogo Almeida, Carroll Wainwright, Pamela Mishkin, Chong Zhang, Sandhini Agarwal, Katarina Slama, Alex Ray, et~al.
\newblock Training language models to follow instructions with human feedback.
\newblock {\em Advances in Neural Information Processing Systems (NeurIPS)}, 35:27730--27744, 2022.

\bibitem{min2023autonomous}
Jihong Min, Stepan Demchyshyn, Juliane~R Sempionatto, Yu~Song, Bekele Hailegnaw, Changhao Xu, Yiran Yang, Samuel Solomon, Christoph Putz, Lukas~E Lehner, et~al.
\newblock An autonomous wearable biosensor powered by a perovskite solar cell.
\newblock {\em Nature Electronics}, pages 1--12, 2023.

\bibitem{bohr2017cmos}
Mark~T Bohr and Ian~A Young.
\newblock Cmos scaling trends and beyond.
\newblock {\em IEEE Micro}, 37(6):20--29, 2017.

\bibitem{taur2002cmos}
Yuan Taur.
\newblock Cmos design near the limit of scaling.
\newblock {\em IBM Journal of Research and Development}, 46(2.3):213--222, 2002.

\bibitem{he2016deep}
Kaiming He, Xiangyu Zhang, Shaoqing Ren, and Jian Sun.
\newblock Deep residual learning for image recognition.
\newblock In {\em Proceedings of the IEEE conference on computer vision and pattern recognition (CVPR)}, pages 770--778. IEEE, 2016.

\bibitem{rodgers1985improvements}
David~P Rodgers.
\newblock Improvements in multiprocessor system design.
\newblock {\em ACM SIGARCH Computer Architecture News}, 13(3):225--231, 1985.

\bibitem{hiramoto2019five}
Toshiro Hiramoto.
\newblock Five nanometre cmos technology.
\newblock {\em Nature Electronics}, 2(12):557--558, 2019.

\bibitem{salahuddin2018era}
Sayeef Salahuddin, Kai Ni, and Suman Datta.
\newblock The era of hyper-scaling in electronics.
\newblock {\em Nature Electronics}, 1(8):442--450, 2018.

\bibitem{sklansky1960conditional}
Jack Sklansky.
\newblock Conditional-sum addition logic.
\newblock {\em IRE Transactions on Electronic computers}, pages 226--231, 1960.

\bibitem{wallace1964suggestion}
Christopher~S Wallace.
\newblock A suggestion for a fast multiplier.
\newblock {\em IEEE Transactions on electronic Computers}, pages 14--17, 1964.

\bibitem{roy2013towards}
Subhendu Roy, Mihir Choudhury, Ruchir Puri, and David~Z Pan.
\newblock Towards optimal performance-area trade-off in adders by synthesis of parallel prefix structures.
\newblock In {\em Proceedings of the Annual DAC}, pages 1--8. IEEE, 2013.

\bibitem{roy2014towards}
Subhendu Roy, Mihir Choudhury, Ruchir Puri, and David~Z Pan.
\newblock Towards optimal performance-area trade-off in adders by synthesis of parallel prefix structures.
\newblock {\em IEEE Transactions on Computer-Aided Design of Integrated Circuits and Systems (TCAD)}, 33(10):1517, 2014.

\bibitem{xiao2021gomil}
Weihua Xiao, Weikang Qian, and Weiqiang Liu.
\newblock Gomil: Global optimization of multiplier by integer linear programming.
\newblock In {\em Design, Automation \& Test in Europe Conference \& Exhibition (DATE)}, pages 374--379. IEEE, 2021.

\bibitem{roy2021prefixrl}
Rajarshi Roy, Jonathan Raiman, Neel Kant, Ilyas Elkin, Robert Kirby, Michael Siu, Stuart Oberman, Saad Godil, and Bryan Catanzaro.
\newblock Prefixrl: Optimization of parallel prefix circuits using deep reinforcement learning.
\newblock In {\em Proceedings of the Annual Design Automation Conference (DAC)}, pages 853--858. IEEE, 2021.

\bibitem{dongsheng2023rl-mul}
Dongsheng Zuo, Yikang Ouyang, and Yuzhe Ma.
\newblock {RL-MUL}: Multiplier design optimization with deep reinforcement learning.
\newblock In {\em Proceedings of the Annual Design Automation Conference (DAC)}, pages 1--8. IEEE, 2023.

\bibitem{geng2021high}
Hao Geng, Yuzhe Ma, Qi~Xu, Jin Miao, Subhendu Roy, and Bei Yu.
\newblock High-speed adder design space exploration via graph neural processes.
\newblock {\em IEEE Transactions on Computer-Aided Design of Integrated Circuits and Systems (TCAD)}, 41(8):2657--2670, 2021.

\bibitem{ma2018cross}
Yuzhe Ma, Subhendu Roy, Jin Miao, Jiamin Chen, and Bei Yu.
\newblock Cross-layer optimization for high speed adders: A pareto driven machine learning approach.
\newblock {\em IEEE Transactions on Computer-Aided Design of Integrated Circuits and Systems (TCAD)}, 38(12):2298--2311, 2018.

\bibitem{silver2016mastering}
David Silver, Aja Huang, Chris~J Maddison, Arthur Guez, Laurent Sifre, George Van Den~Driessche, Julian Schrittwieser, Ioannis Antonoglou, Veda Panneershelvam, Marc Lanctot, et~al.
\newblock Mastering the game of go with deep neural networks and tree search.
\newblock {\em nature}, 529(7587):484--489, 2016.

\bibitem{palnitkar2003verilog}
Samir Palnitkar.
\newblock {\em Verilog HDL: a guide to digital design and synthesis}, volume~1.
\newblock Prentice Hall Professional, 2003.

\bibitem{browne2012survey}
Cameron~B Browne, Edward Powley, Daniel Whitehouse, Simon~M Lucas, Peter~I Cowling, Philipp Rohlfshagen, Stephen Tavener, Diego Perez, Spyridon Samothrakis, and Simon Colton.
\newblock A survey of monte carlo tree search methods.
\newblock {\em IEEE Transactions on Computational Intelligence and AI in games}, 4(1):1--43, 2012.

\bibitem{schulman2017proximal}
John Schulman, Filip Wolski, Prafulla Dhariwal, Alec Radford, and Oleg Klimov.
\newblock Proximal policy optimization algorithms.
\newblock {\em arXiv preprint arXiv:1707.06347}, 2017.

\bibitem{kogge1973parallel}
Peter~M Kogge and Harold~S Stone.
\newblock A parallel algorithm for the efficient solution of a general class of recurrence equations.
\newblock {\em IEEE Transactions on Computers (TC)}, 100(8):786--793, 1973.

\bibitem{ladner1980parallel}
Richard~E Ladner and Michael~J Fischer.
\newblock Parallel prefix computation.
\newblock {\em Journal of the ACM (JACM)}, 27(4):831--838, 1980.

\bibitem{roth2020fundamentals}
Charles~H Roth~Jr, Larry~L Kinney, and Eugene~B John.
\newblock {\em Fundamentals of logic design}.
\newblock Cengage Learning, 2020.

\bibitem{swiechowski2023monte}
Maciej {\'S}wiechowski, Konrad Godlewski, Bartosz Sawicki, and Jacek Ma{\'n}dziuk.
\newblock Monte carlo tree search: A review of recent modifications and applications.
\newblock {\em Artificial Intelligence Review}, 56(3):2497--2562, 2023.

\bibitem{vinyals2019grandmaster}
Oriol Vinyals, Igor Babuschkin, Wojciech~M Czarnecki, Micha{\"e}l Mathieu, Andrew Dudzik, Junyoung Chung, David~H Choi, Richard Powell, Timo Ewalds, Petko Georgiev, et~al.
\newblock Grandmaster level in starcraft ii using multi-agent reinforcement learning.
\newblock {\em Nature}, 575(7782):350--354, 2019.

\bibitem{fawzi2022discovering}
Alhussein Fawzi, Matej Balog, Aja Huang, Thomas Hubert, Bernardino Romera-Paredes, Mohammadamin Barekatain, Alexander Novikov, Francisco~J R~Ruiz, Julian Schrittwieser, Grzegorz Swirszcz, et~al.
\newblock Discovering faster matrix multiplication algorithms with reinforcement learning.
\newblock {\em Nature}, 610(7930):47--53, 2022.

\bibitem{kocsis2006bandit}
Levente Kocsis and Csaba Szepesv{\'a}ri.
\newblock Bandit based monte-carlo planning.
\newblock In {\em European conference on machine learning}, pages 282--293. Springer, 2006.

\bibitem{dedieu2017deep}
Antoine Dedieu and Jonathan Amar.
\newblock Deep reinforcement learning for 2048.
\newblock In {\em Conference on Neural Information Processing Systems (NeurIPS)}, 2017.

\bibitem{murtagh1991multilayer}
Fionn Murtagh.
\newblock Multilayer perceptrons for classification and regression.
\newblock {\em Neurocomputing}, 2(5-6):183--197, 1991.

\bibitem{bishop2006pattern}
Christopher~M Bishop and Nasser~M Nasrabadi.
\newblock {\em Pattern recognition and machine learning}, volume~4.
\newblock Springer, 2006.

\bibitem{girshick2015fast}
Ross Girshick.
\newblock Fast r-cnn.
\newblock In {\em Proceedings of the IEEE international conference on computer vision (ICCV)}, pages 1440--1448. IEEE, 2015.

\bibitem{zhang2022easymac}
Jiaxi Zhang, Qiuyang Gao, Yijiang Guo, Bizhao Shi, and Guojie Luo.
\newblock Easymac: design exploration-enabled multiplier-accumulator generator using a canonical architectural representation.
\newblock In {\em Proceedings of Asia and South Pacific Design Automation Conference (ASP-DAC)}, pages 647--653. IEEE, 2022.

\bibitem{wolf2016yosys}
Clifford Wolf.
\newblock Yosys open synthesis suite, 2016.

\bibitem{ajayi2019openroad}
Tutu Ajayi and David Blaauw.
\newblock Openroad: Toward a self-driving, open-source digital layout implementation tool chain.
\newblock In {\em Proceedings of Government Microcircuit Applications and Critical Technology Conference}, 2019.

\bibitem{nangate2008nangate}
Inc NanGate.
\newblock {NanGate FreePDK45} open cell library, 2008.

\bibitem{clark2016asap7}
Lawrence~T Clark, Vinay Vashishtha, Lucian Shifren, Aditya Gujja, Saurabh Sinha, Brian Cline, Chandarasekaran Ramamurthy, and Greg Yeric.
\newblock {ASAP7}: A 7-nm {FinFET} predictive process design kit.
\newblock {\em Microelectronics Journal}, 53, 2016.

\bibitem{snir1986depth}
Marc Snir.
\newblock Depth-size trade-offs for parallel prefix computation.
\newblock {\em Journal of Algorithms}, 7(2):185--201, 1986.

\bibitem{matsunaga2007area}
Taeko Matsunaga and Yusuke Matsunaga.
\newblock Area minimization algorithm for parallel prefix adders under bitwise delay constraints.
\newblock In {\em Proceedings of the 17th ACM Great Lakes symposium on VLSI}, pages 435--440, 2007.

\bibitem{brent1982regular}
Brent and Kung.
\newblock A regular layout for parallel adders.
\newblock {\em IEEE Transactions on Computers (TC)}, 100(3):260--264, 1982.

\bibitem{behrooz2000computer}
Parhami Behrooz.
\newblock Computer arithmetic: Algorithms and hardware designs.
\newblock {\em Oxford University Press}, 19:512583--512585, 2000.

\bibitem{van1987simulated}
Peter~JM Van~Laarhoven, Emile~HL Aarts, Peter~JM van Laarhoven, and Emile~HL Aarts.
\newblock {\em Simulated annealing}.
\newblock Springer, 1987.

\bibitem{ousterhout1989tcl}
John~K Ousterhout et~al.
\newblock {\em Tcl: An embeddable command language}.
\newblock University of California, Berkeley, Computer Science Division, 1989.

\bibitem{compiler2016synopsys}
Synopsys~Design Compiler.
\newblock Synopsys design compiler.
\newblock {\em Pages/default. aspx}, 2016.

\bibitem{weste2015cmos}
Neil~HE Weste and David Harris.
\newblock {\em CMOS VLSI design: a circuits and systems perspective}.
\newblock Pearson Education India, 2015.

\bibitem{han1987vlsi}
Tackdon Han, David~A Carlson, and Steven~P Levitan.
\newblock {\em VLSI DESIGN OF HIGH-SPEED, LOW-AREA ADDITION CIRCUITRY.}
\newblock IEEE, 1987.

\bibitem{liu2003algorithmic}
Jianhua Liu, Shuo Zhou, Haikun Zhu, and Chung-Kuan Cheng.
\newblock An algorithmic approach for generic parallel adders.
\newblock In {\em International Conference on Computer Aided Design (ICCAD)}, pages 734--740. IEEE, 2003.

\bibitem{fishburn1991depth}
John~P Fishburn.
\newblock A depth-decreasing heuristic for combinational logic: or how to convert a ripple-carry adder into a carry-lookahead adder or anything in-between.
\newblock In {\em ACM/IEEE DAC}, pages 361--364. ACM/IEEE, 1991.

\bibitem{zimmermann1996non}
Reto Zimmermann.
\newblock Non-heuristic optimization and synthesis of parallel-prefix adders.
\newblock In {\em proc. of IFIP workshop}. Citeseer, 1996.

\bibitem{zhu2005constructing}
Haikun Zhu, Chung-Kuan Cheng, and Ronald Graham.
\newblock Constructing zero-deficiency parallel prefix adder of minimum depth.
\newblock In {\em Proceedings of Asia and South Pacific Design Automation Conference (ASP-DAC)}, pages 883--888, 2005.

\bibitem{melo2001convergence}
Francisco~S Melo.
\newblock Convergence of q-learning: A simple proof.
\newblock {\em Institute Of Systems and Robotics, Tech. Rep}, pages 1--4, 2001.

\bibitem{dadda1965some}
Luigi Dadda.
\newblock Some schemes for parallel multipliers.
\newblock {\em Alta frequenza}, 34:349--356, 1965.

\bibitem{townsend2003comparison}
Whitney~J Townsend, Earl~E Swartzlander~Jr, and Jacob~A Abraham.
\newblock A comparison of dadda and wallace multiplier delays.
\newblock In {\em Advanced signal processing algorithms, architectures, and implementations XIII}, volume 5205, pages 552--560. SPIE, 2003.

\bibitem{mankowitz2023faster}
Daniel~J Mankowitz, Andrea Michi, Anton Zhernov, Marco Gelmi, Marco Selvi, Cosmin Paduraru, Edouard Leurent, Shariq Iqbal, Jean-Baptiste Lespiau, Alex Ahern, et~al.
\newblock Faster sorting algorithms discovered using deep reinforcement learning.
\newblock {\em Nature}, 618(7964):257--263, 2023.

\bibitem{liu2022learn}
Jinxin Liu, Donglin Wang, Qiangxing Tian, and Zhengyu Chen.
\newblock Learn goal-conditioned policy with intrinsic motivation for deep reinforcement learning.
\newblock In {\em Proceedings of the AAAI conference on artificial intelligence}, 2022.

\bibitem{liu2021unsupervised}
Jinxin Liu, Hao Shen, Donglin Wang, Yachen Kang, and Qiangxing Tian.
\newblock Unsupervised domain adaptation with dynamics-aware rewards in reinforcement learning.
\newblock {\em NeurIPS}, 34:28784--28797, 2021.

\bibitem{mnih2013playing}
Volodymyr Mnih, Koray Kavukcuoglu, David Silver, Alex Graves, Ioannis Antonoglou, Daan Wierstra, and Martin Riedmiller.
\newblock Playing atari with deep reinforcement learning.
\newblock {\em arXiv preprint arXiv:1312.5602}, 2013.

\bibitem{lillicrap2015continuous}
Timothy~P Lillicrap, Jonathan~J Hunt, Alexander Pritzel, Nicolas Heess, Tom Erez, Yuval Tassa, David Silver, and Daan Wierstra.
\newblock Continuous control with deep reinforcement learning.
\newblock {\em arXiv preprint arXiv:1509.02971}, 2015.

\bibitem{sutton1999policy}
Richard~S Sutton, David McAllester, Satinder Singh, and Yishay Mansour.
\newblock Policy gradient methods for reinforcement learning with function approximation.
\newblock {\em Conference on Neural Information Processing Systems (NeurIPS)}, 12, 1999.

\bibitem{zhuang2023behavior}
Zifeng Zhuang, Kun Lei, Jinxin Liu, Donglin Wang, and Yilang Guo.
\newblock Behavior proximal policy optimization.
\newblock {\em arXiv preprint arXiv:2302.11312}, 2023.

\bibitem{liu2022dara}
Jinxin Liu, Hongyin Zhang, and Donglin Wang.
\newblock Dara: Dynamics-aware reward augmentation in offline reinforcement learning.
\newblock {\em arXiv preprint arXiv:2203.06662}, 2022.

\bibitem{liu2024ceil}
Jinxin Liu, Li~He, Yachen Kang, Zifeng Zhuang, Donglin Wang, and Huazhe Xu.
\newblock Ceil: Generalized contextual imitation learning.
\newblock {\em NeurIPS}, 36, 2024.

\bibitem{hosny2020drills}
Abdelrahman Hosny, Soheil Hashemi, Mohamed Shalan, and Sherief Reda.
\newblock Drills: Deep reinforcement learning for logic synthesis.
\newblock In {\em 2020 25th Asia and South Pacific Design Automation Conference (ASP-DAC)}, pages 581--586. IEEE, 2020.

\bibitem{zhu2020exploring}
Keren Zhu, Mingjie Liu, Hao Chen, Zheng Zhao, and David~Z Pan.
\newblock Exploring logic optimizations with reinforcement learning and graph convolutional network.
\newblock In {\em Proceedings of the 2020 ACM/IEEE Workshop on Machine Learning for CAD}, pages 145--150, 2020.

\bibitem{niu2023ossp}
Dan Niu, Yichao Dong, Zhou Jin, Chuan Zhang, Qi~Li, and Changyin Sun.
\newblock Ossp-pta: An online stochastic stepping policy for pta on reinforcement learning.
\newblock {\em IEEE Transactions on Computer-Aided Design of Integrated Circuits and Systems}, 2023.

\bibitem{jin2022accelerating}
Zhou Jin, Haojie Pei, Yichao Dong, Xiang Jin, Xiao Wu, Wei~W Xing, and Dan Niu.
\newblock Accelerating nonlinear dc circuit simulation with reinforcement learning.
\newblock In {\em Proceedings of the 59th ACM/IEEE Design Automation Conference}, pages 619--624, 2022.

\bibitem{mirhoseini2021graph}
Azalia Mirhoseini, Anna Goldie, Mustafa Yazgan, Joe~Wenjie Jiang, Ebrahim Songhori, Shen Wang, Young-Joon Lee, Eric Johnson, Omkar Pathak, Azade Nazi, et~al.
\newblock A graph placement methodology for fast chip design.
\newblock {\em Nature}, 594(7862):207--212, 2021.

\bibitem{cheng2021joint}
Ruoyu Cheng and Junchi Yan.
\newblock On joint learning for solving placement and routing in chip design.
\newblock {\em Advances in Neural Information Processing Systems (NeurIPS)}, 34:16508--16519, 2021.

\bibitem{lai2022maskplace}
Yao Lai, Yao Mu, and Ping Luo.
\newblock Maskplace: Fast chip placement via reinforced visual representation learning.
\newblock {\em Advances in Neural Information Processing Systems (NeurIPS)}, 35:24019--24030, 2022.

\bibitem{shi2024macro}
Yunqi Shi, Ke~Xue, Song Lei, and Chao Qian.
\newblock Macro placement by wire-mask-guided black-box optimization.
\newblock {\em Advances in Neural Information Processing Systems (NeurIPS)}, 36, 2024.

\bibitem{lai2023chipformer}
Yao Lai, Jinxin Liu, Zhentao Tang, Bin Wang, Jianye Hao, and Ping Luo.
\newblock Chipformer: Transferable chip placement via offline decision transformer.
\newblock In {\em ICML}, pages 18346--18364. PMLR, 2023.

\bibitem{zhong2024preroutgnn}
Ruizhe Zhong, Junjie Ye, Zhentao Tang, Shixiong Kai, Mingxuan Yuan, Jianye Hao, and Junchi Yan.
\newblock Preroutgnn for timing prediction with order preserving partition: Global circuit pre-training, local delay learning and attentional cell modeling.
\newblock {\em arXiv preprint arXiv:2403.00012}, 2024.

\bibitem{chen2023reinforcement}
Hao Chen, Kai-Chieh Hsu, Walker~J Turner, Po-Hsuan Wei, Keren Zhu, David~Z Pan, and Haoxing Ren.
\newblock Reinforcement learning guided detailed routing for custom circuits.
\newblock In {\em Proceedings of the 2023 International Symposium on Physical Design (ISPD)}, pages 26--34, 2023.

\bibitem{qu2021asynchronous}
Tong Qu, Yibo Lin, Zongqing Lu, Yajuan Su, and Yayi Wei.
\newblock Asynchronous reinforcement learning framework for net order exploration in detailed routing.
\newblock In {\em 2021 Design, Automation \& Test in Europe Conference \& Exhibition (DATE)}, pages 1815--1820. IEEE, 2021.

\bibitem{beheshti2021reinforced}
Sayed~Aresh Beheshti-Shirazi, Ashkan Vakil, Sai Manoj, Ioannis Savidis, Houman Homayoun, and Avesta Sasan.
\newblock A reinforced learning solution for clock skew engineering to reduce peak current and ir drop.
\newblock In {\em Proceedings of the 2021 on Great Lakes Symposium on VLSI}, pages 181--187, 2021.

\bibitem{wang2020gcn}
Hanrui Wang, Kuan Wang, Jiacheng Yang, Linxiao Shen, Nan Sun, Hae-Seung Lee, and Song Han.
\newblock Gcn-rl circuit designer: Transferable transistor sizing with graph neural networks and reinforcement learning.
\newblock In {\em 2020 57th ACM/IEEE Design Automation Conference (DAC)}, pages 1--6. IEEE, 2020.

\bibitem{lu2021rl}
Yi-Chen Lu, Siddhartha Nath, Vishal Khandelwal, and Sung~Kyu Lim.
\newblock Rl-sizer: Vlsi gate sizing for timing optimization using deep reinforcement learning.
\newblock In {\em 2021 58th ACM/IEEE DAC}, pages 733--738. IEEE, 2021.

\bibitem{shi2022deeptpi}
Zhengyuan Shi, Min Li, Sadaf Khan, Liuzheng Wang, Naixing Wang, Yu~Huang, and Qiang Xu.
\newblock Deeptpi: Test point insertion with deep reinforcement learning.
\newblock In {\em 2022 IEEE International Test Conference (ITC)}, pages 194--203. IEEE, 2022.

\end{thebibliography}

\clearpage
\appendix

\section{Method Details}
\label{method_detail}

\subsection{Level upper bound}
\begin{figure}[!ht]%
\centering
\includegraphics[width=0.6\linewidth]{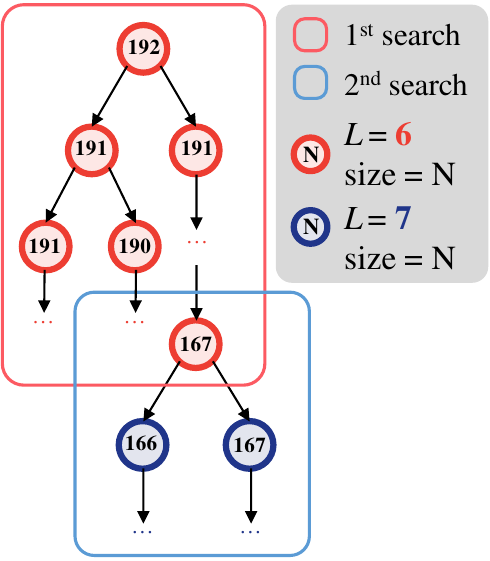}
\caption{Level upper bound \textit{L} for optimizing theoretical metrics of adders.
The example is for 64-bit adder design.
The search is divided into stages, and the level upper bound $L$ increases one at a time.
The initial state for each search is set to the best adder found in the last search iteration.
}\label{level_upper_bound}
\Description{}
\end{figure}

When optimizing adders in theoretical metrics, the search process is stratified based on a series of incremental level upper bounds, $L$. 
The initial bound is set to $\log_2 N$ and is incrementally raised in subsequent stages. 
For each new stage, the starting configuration state is the adder design with the minimum size obtained from the previous stage's search, as illustrated in Fig. \ref{level_upper_bound}.

\subsection{Two-level retrieval}
\begin{figure}[!ht]%
\centering
\includegraphics[width=0.6\linewidth]{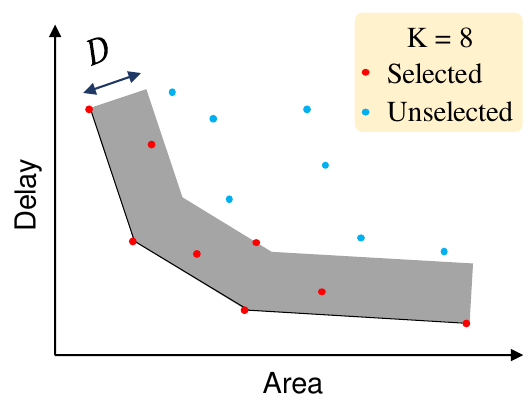}
\caption{\textbf{Select adders in two-level retrieval.} 
After the first stage of the two-level retrieval process, each adder is represented by a 2D point based on its delay and area. When selecting the top \textit{K} adders for the second stage, we sort the adders according to their distances from these points to the Pareto boundary. The \textit{K} adders with the smallest distances are selected. The threshold distance, denoted as \textit{D}, is defined by the distance of the \textit{K}-th adder to the Pareto boundary.
}\label{pareto_topk}
\Description{}
\end{figure}

In our two-level retrieval strategy, we implement a fast synthesis flow with minimal loss of precision. 
In the fast flow, we keep all other steps, including logic synthesis, clock tree synthesis, and placement, but remove the most expensive routing step. 
According to our efficiency test in Fig. \ref{efficiency}, our fast synthesis flow without the routing step can speed up more than ten times.
At the same time, the fast flow can still achieve highly accurate area measurements and 95\% accurate delay estimations as detailed in Appendix Table~\ref{simplified_synthesis}.
Thus, the fast synthesis flow can help search for as many adders as possible without losing accuracy.
At the end of the first stage of two-level retrieval, we use the coordinate $(\text{area}, \text{delay})$ as the representative points for adders and compute all distances from these points to the Pareto boundary. 
We sort the distances in ascending order and use the $K$-th distance $D$ as the threshold for selecting the adders to the second stage.
As shown in Fig. \ref{pareto_topk}, the $K$ adders with the shortest distances to the Pareto boundary—constituting the top 10\% in our efficiency settings—will be selected for full synthesis execution.

\subsection{Synthesis script}
\label{synthesis_script}
\begin{figure}[!htbp]%
\centering
\includegraphics[width=0.8\linewidth]{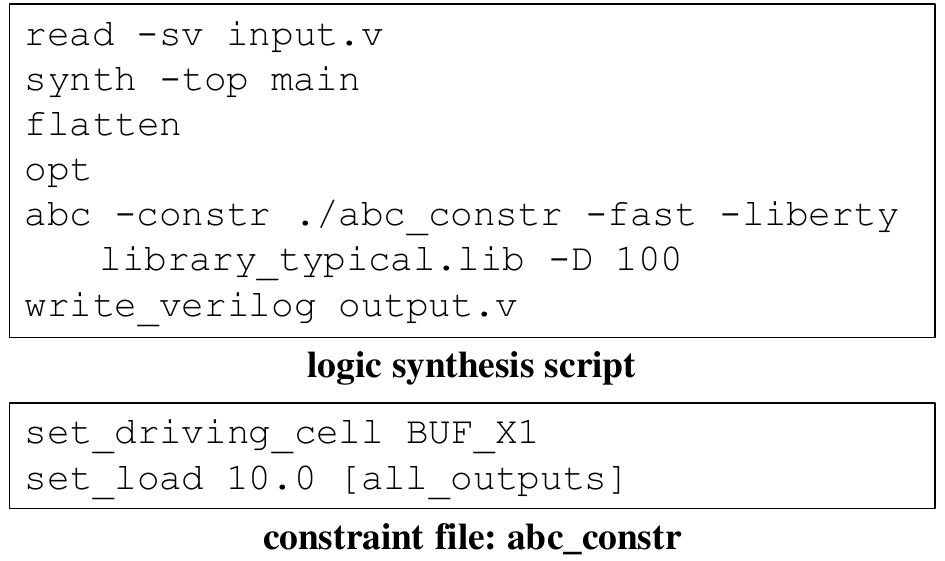}
\caption{Scripts for logical synthesis.
}\label{yosys_script}
\Description{}
\end{figure}

\begin{figure*}[!thbp]%
\centering
\includegraphics[width=0.9\linewidth]{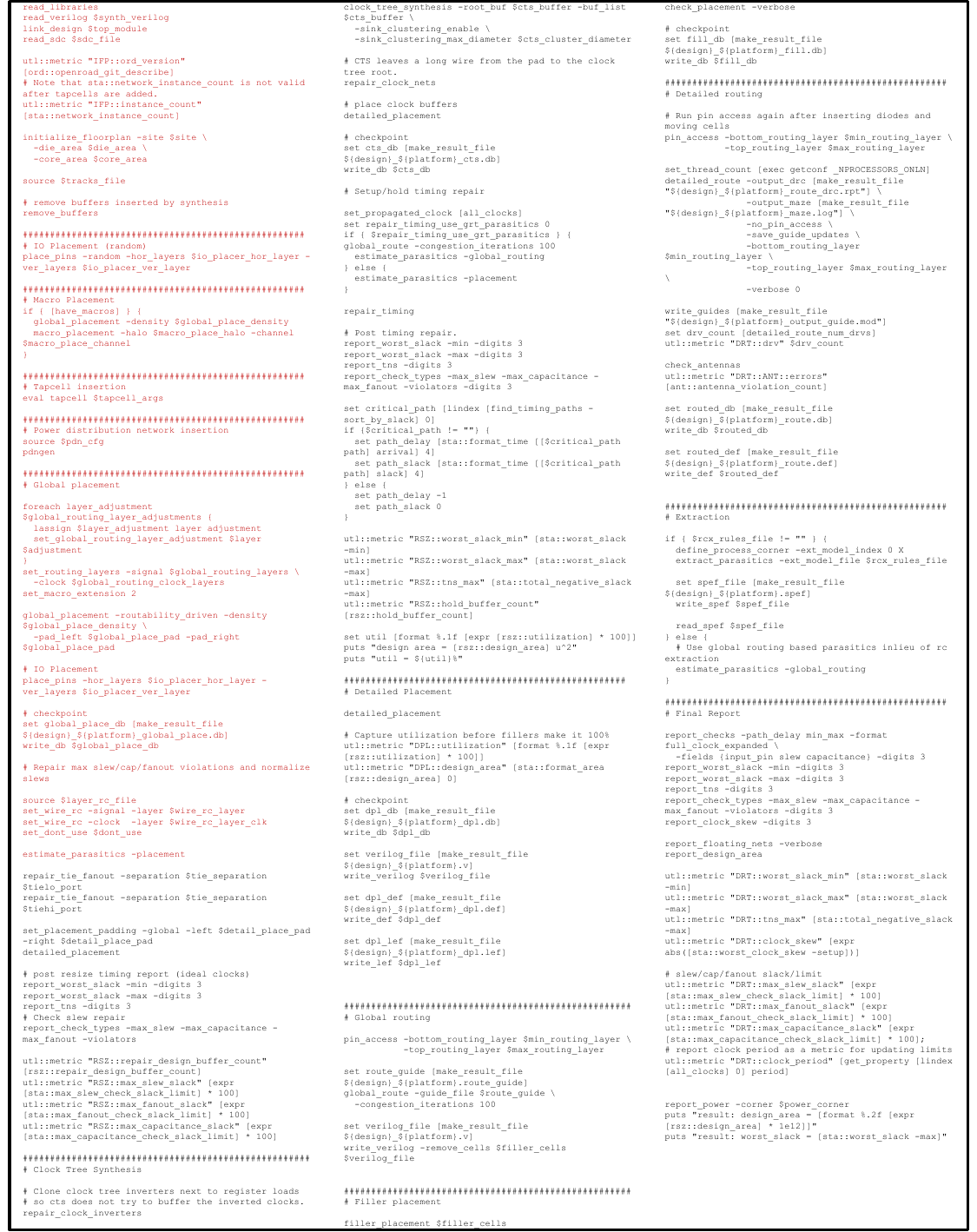}
\caption{\textbf{Scripts for physical synthesis.} 
\textcolor{BrickRed}{Red} parts of scripts are the fast synthesis flow in our two-level retrieval strategy, which excludes the most time-consuming routing phase.
}\label{openroad_script}
\Description{}
\end{figure*}

The process of logical and physical synthesis for Yosys and OpenROAD is implemented using the Tcl scripting language \cite{ousterhout1989tcl}. We provide the complete Tcl scripts that we used for the logical synthesis in Fig. \ref{yosys_script} and for the physical synthesis in Fig. \ref{openroad_script}.

\subsection{Caching and design save/recover}
According to our configuration, both the prefix and compressor trees can be easily saved and subsequently restored. The prefix tree is stored as an upper triangular matrix $A_{N \times N}$, where a cell $(i,j)$ is marked with $A_{i,j} = 1$ if it exists; otherwise, $A_{i,j} = 0$ if it does not.
The compressor tree is represented by a variable-length sequence $S = \{a_0, a_1, a_2, \ldots, a_{T-1}\}$ with each $a_i \in {0, 1}$. Here, $a_i = 0$ represents the addition of a full adder, and $a_i = 1$ signifies the addition of a half adder.
In the context of our game modeling, the matrix $A$ and the sequence $S$ together can completely reconstruct the prefix and compressor trees, respectively. Furthermore, both structures can be serialized into strings. These strings are then processed through a hash function to generate fixed-length values that serve as keys in our cache. This cache stores the results of previous syntheses, which helps in avoiding redundant synthesis runs.
\looseness=-1

\subsection{Hyperparameter}
\label{hyperparameter}
Our hyperparameter configuration can be found in Table \ref{Hyperparameter}.

\begin{table}[!ht]
	\caption{Hyperparameter Configuration}
	\label{Hyperparameter}
    \centering
    \small
    \begin{threeparttable}
    \begin{tabular}{c|c|c}
    \toprule
        Name & Description & Value\\ \midrule
        $\alpha$ & Weight for delay and area in Fig. \ref{mcts} & $0.01/0.001/0$ \tnote{*} \\
        $\beta$ & Sum weight in Eq. \ref{action_value} & $0.01$\\
       $c$ & Sum weight in Eq. \ref{value} & $10\sqrt{2}$\\
       $p$ & Penalty value for using half adders & $0.1$ \\
       $\gamma$ & Discount factor in Eq. \ref{loss_func} & $0.8$ \\
       $\epsilon$ & Gradient clip norm in Eq. \ref{loss} & $0.2$\\
       - & Batch size for PPO & $64$\\
        - & Replay buffer size for PPO & $6N^2$\\
        - & learning rate & $0.001$\\ 
        \bottomrule
    \end{tabular}
    \begin{tablenotes}
    
    \footnotesize
    \item[*] $0.01$ for designing multipliers, $0$ (ripple-carry adder as initial state) and $0.001$ (others) for designing adders, where the unit of delay is $ns$, and the unit of area is $\mu m^2$. 
    \end{tablenotes}
    \end{threeparttable}
\end{table}

\subsection{State features}
Details of state features for policy and value networks as inputs are shown in Table \ref{tb_feature}. All features are normalized before inputting into the networks.

\begin{table}[!ht]
\caption{State features for policy and value network}\label{tb_feature}%
\resizebox{\linewidth}{!}{
\begin{tabular}{@{}l|c|l@{}}
\toprule
Feature & Size &Description\\
\midrule
Action digit  & $1$ & Digit for action. \\
Max delay & $1$  & Maximum estimated delay value of all bits.     \\
Number of half adders  & $1$  & Number of added half adders in action digit. \\
Mask for action & $2$ & The mask for ensuring valid action. \\
Delay of action bits & $3$ & The delays of bits for action. \\
\bottomrule
\end{tabular}
}
\end{table}

\subsection{Input selection in compressor tree}
In the compressor tree design, both half adders and full adders are utilized. When assigning input bits to a full or half adder, we prioritize the bits with minimal estimated delays. For instance, consider the case where we are selecting inputs for action $a_0$ and the available input bits have delays $\{0, 0, 0, 0, 1\}$. In this situation, the three bits with a delay of $0$ would be chosen as inputs for a full adder to minimize the overall delay.
The rationale behind this is that adders introduce additional delays, and our objective is to minimize the maximum delay across all bits. A more nuanced strategy is employed when inputs are fed into a full adder: the bit with the highest delay out of the three is connected to the carry input. For example, given input bits with delays $\{0, 0, 1\}$, the bit with a delay of $1$ would be connected to the carry input of the full adder. This strategy is adopted because the delay from the carry input to the output bits involves only two logic gates, which is faster than the three logic gates' delay from the addend inputs to the outputs.

\subsection{Strategy for searching multipliers}
In the search process, each multiplier is tested on two boundary expected delay parameters ($50$ and $2\times 10^5$). The average delay and area are then calculated from the results obtained at these two boundary conditions. The performance score for each multiplier is the weighted sum of the average delay and area. The multiplier with the highest score is selected for final evaluation.

\section{Supplementary Results}

\subsection{Adder design}
\begin{figure*}[!thbp]%
\centering
\includegraphics[width=\linewidth]{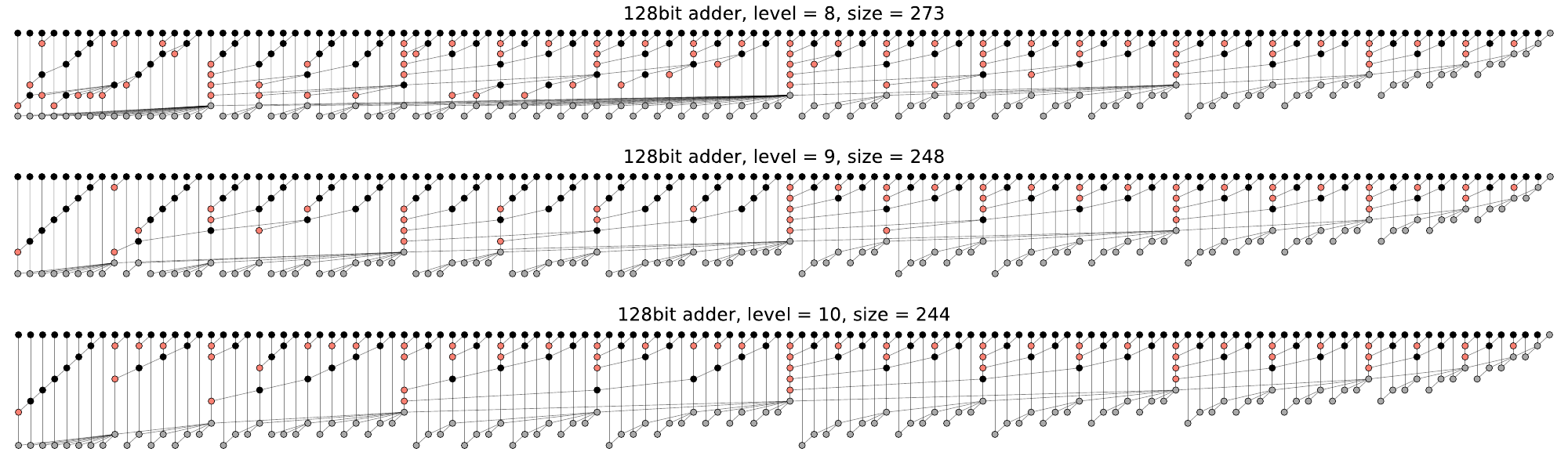}
\caption{\textbf{Additional examples of 128-bit adders.} More structures of the 128-bit adder first discovered by our method are shown.
}
\label{additional_adder}
\Description{}
\end{figure*}

In addition to Fig. \ref{prefix_tree}, we present some novel designs of the $128$-bit adder discovered by our method in Fig. \ref{additional_adder}, which also achieve minimal sizes under the given levels.

\begin{figure*}[!thbp]%
\centering
\includegraphics[width=\linewidth]{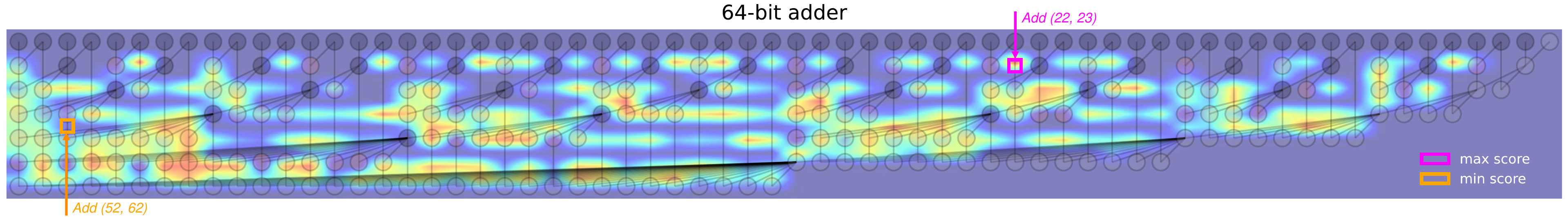}
\caption{Heatmap for first action scores. (cont.)
}
\label{heatmap_supp}
\Description{}
\end{figure*}

Besides the heatmaps in Fig. \ref{heatmap}, we have presented the heatmap for the $64$-bit adder in Fig. \ref{heatmap_supp}. For the $64$-bit adder, the action of adding the cell at position $(22, 23)$, which has the highest score, is selected as the first action primarily because it does not increase the level. Conversely, the action of adding the cell at position $(52, 62)$ scores the lowest, primarily due to the consequent increase in level, which in turn raises the delay.

\subsection{Correlation between metrics}

\begin{figure*}[!th]%
\centering
\includegraphics[width=\linewidth]{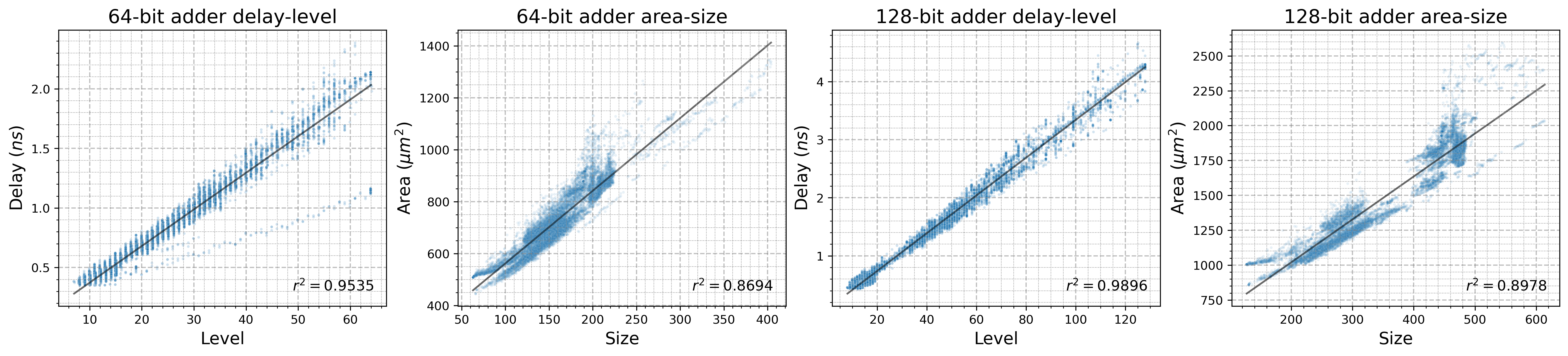}
\caption{Correlation of theoretical and practical metrics.
The fitted lines indicate strong correlations in delay-level and area-size.
The data are derived from 6k adders for each.
}\label{level_delay_scatter}
\Description{}
\end{figure*}

We investigated the correlation between theoretical and practical metrics to demonstrate the significance of optimizing theoretical metrics. For $64$-bit and $128$-bit adders, we sampled $6,000$ instances to assess their theoretical and practical metrics. Our results show that there is a high correlation within two groups of metrics: level with delay, and size with area, as illustrated in Fig. \ref{level_delay_scatter}. Thus, structures with lower levels and smaller sizes are more likely to result in adders with lower delays and smaller areas.

\subsection{Commercial synthesis tool}

In addition to our tests on open-source tools, we also utilized a commercial synthesis tool, Synopsys Design Compiler 2020 \cite{compiler2016synopsys}, to demonstrate the generalizability of our approach. Table \ref{commercial} presents the results of the multipliers designed by this tool. We did not incorporate timing constraints when testing the delay of the critical path. The technology library used was the Nangate 45nm library~\cite{nangate2008nangate}. 
The speed of our designed multiplier still holds a significant advantage, illustrating our design approach's broad applicability and substantial potential.

\begin{table*}[!htbp]
\caption{\textbf{Results of a commercial synthesis tool.} All designs are the best-discovered multipliers with the OpenROAD tool. Corresponding Verilog codes are input into the Synopsis Design Compiler for synthesis. 
}\label{commercial}%
\centering
\small
\begin{tabular}{l|cc|cc|cc|cc}
\toprule
Bits in multiplier & \multicolumn{2}{c|}{8-bit}& \multicolumn{2}{c|}{16-bit} & \multicolumn{2}{c|}{32-bit} & \multicolumn{2}{c}{64-bit} \\
\midrule
Method  & area ($\mu m^2$) & delay ($ns$) & area ($\mu m^2$) & delay  ($ns$)& area ($\mu m^2$) & delay ($ns$) & area ($\mu m^2$)& delay ($ns$)\\
\midrule
Default           & 314.1 & 1.30       & 1288.5         & 2.60       & 5203.2         & 4.88       & 20844.3        & 9.29       \\
RL-MUL            & 313.9          & 1.47       & 1373.6         & 2.98       & 5757.3         & 5.86       & 23563.6        & 11.73      \\
MCTS        & 416.6          & 1.65       & 1734.9         & 3.19       & 7331.5         & 6.07       & 29545.7        & 11.92      \\
PPO+MCTS    & 465.5          & \textbf{1.20} & 1866.8 & \textbf{1.76} & 7555.5 & \textbf{2.34} & 30134.1 & \textbf{3.17}  \\
\bottomrule
\end{tabular}
\end{table*}

\subsection{Design time}
The overall design time is reported in Table \ref{time}, and the duration is within an acceptable range for the design process.

\begin{table}[ht]
\caption{Total design time}\label{time}%
\small
\begin{tabular}{l|ccc|cccc}
\toprule
Module & \multicolumn{3}{c|}{Adder} & \multicolumn{4}{c}{Multiplier} \\
\midrule
Bits & 32 & 64 & 128 & 8 & 16 & 32 & 64 \\
Time (h) & 1.71 & 3.68 & 7.34 & 0.82  & 2.26 & 4.04 & 27.92\\
\bottomrule
\end{tabular}
\end{table}

\subsection{Simulation accuracy of fast flow}
The time-consuming routing phase is removed in the fast flow of the two-level strategy. To evaluate the impact of this simplification, we tested the simulation accuracy of the fast flow against the full flow. The results in Table \ref{simplified_synthesis} indicate that the fast flow can still achieve an utterly accurate area estimation and over 95\% accurate delay. Therefore, the fast synthesis flow can help improve efficiency without a significant loss of accuracy.

\begin{table}[!ht]
\caption{Accuracy of fast synthesis flow}
\label{simplified_synthesis}
\small
\begin{tabular}{@{}l|c|c|c@{}}
\toprule
Bits of adders & 32 & 64 & 128\\
\midrule
Delay Acc. (\%) & 96.11$\pm$0.86 & 95.82$\pm$1.12 & 95.34$\pm$2.60\\
Area Acc. (\%) & 100.00$\pm$0.00 & 100.00$\pm$0.00 & 100.00$\pm$0.00\\
\bottomrule
\end{tabular}
\end{table}

\section{Related work}

\subsection{Computer arithmetic}

In the quest for high performance and low cost, computer arithmetic design plays a crucial role in computer hardware, one of the most fundamental fields in computer science \cite{behrooz2000computer}. Issues for study include number representation, arithmetic operations, and real arithmetic. The addition is the most common arithmetic operation and serves as a basic unit for many other operations, making it the most studied module. The most basic adder structure is the ripple carry adder, which propagates the carry bit from low to high bits. Due to its serial structure, both the delay and size are $O(N)$ for an $N$-bit addition. The carry look-ahead adder has been proposed to improve the delay by computing the carries for each digit simultaneously through an expanded formula. It can achieve $O(\log N)$ delay and $O(N\log N)$ size. However, due to the long internal delay of higher-valency gates used in the look-ahead adder \cite{weste2015cmos}, various prefix adders have been developed, including the Brent-Kung \cite{brent1982regular}, Sklansky \cite{sklansky1960conditional}, Kogge-Stone \cite{kogge1973parallel}, and Han-Carlson adders \cite{han1987vlsi}. Most of these designs are variations of prefix adders. Although the minimal delay complexity is still $O(\log N)$, these adders can often have lower delays than the carry look-ahead adder because they use faster two-input logic gates \cite{weste2015cmos}. Additionally, different prefix adders can strike a balance between delay and area, making them more suitable for actual hardware design.

Despite extensive research, human-engineered prefix adders continue to encounter challenges in realizing Pareto-optimal designs. Notably, the dimensions of the Sklansky adder can be further minimized whilst maintaining its operational level, as indicated by Roy et al. \cite{roy2013towards}. Consequently, a plethora of optimization-oriented methodologies have been put forward \cite{matsunaga2007area, liu2003algorithmic, fishburn1991depth, zimmermann1996non, snir1986depth, zhu2005constructing}. The heuristic algorithm proposed by Roy and colleagues \cite{roy2013towards} employs a bottom-up enumeration tactic, commencing with a binary adder and iteratively escalating the bit count inductively based on extant structures. To reconcile the disparity between theoretical and empirical metrics, Ma et al. \cite{ma2018cross} developed a training regimen for a predictive model to estimate actual metrics from theoretical ones. This model utilizes a Pareto active learning approach to selectively scrutinize adders, which exhibit latent high-performance metrics, for empirical validation via synthesis tools. Additionally, Geng et al. \cite{geng2021high} have embraced graph neural networks to enhance the precision of the predictive model. However, these methodologies necessitate the preselection of a finite set of adders for prediction purposes, representing merely a fraction of the comprehensive feasible space and consequently potentially overlooking superior adder configurations. The foray of reinforcement learning into the domain of adder design was pioneered by Roy et al. \cite{roy2021prefixrl}, integrating a novel approach to address design challenges. Nevertheless, the employed Q-network methodology \cite{melo2001convergence} is deficient in exploration capabilities when applied to expansive problem domains. Moreover, it mandates complete synthesis for each adder design, a process that is prohibitively time-intensive when attempting to sample a vast array of adder configurations, thereby yielding suboptimal solutions.

In analog to adder design, foundational research on multipliers has also been rooted in manual methodologies. An $N$-bit multiplication fundamentally involves the generation of $N$ partial products through the deployment of $N^2$ AND gates, which correspond to each pair of bits to be multiplied \cite{weste2015cmos}. Subsequently, these partial products are accumulated to yield a $2N$-bit result. The most straightforward strategy employs $N$ successive accumulation operations over $N$ clock cycles, utilizing a serial approach that requires solely one adder and one register. Nevertheless, this method incurs a delay of $O(N\log N)$ with the employment of a logarithmic delay adder \cite{behrooz2000computer}, indicating a super-linear increase relative to the bit count. To elevate computational efficiency, one may adopt a compressor tree structure to compress the partial products concurrently using full and half adders, finalizing the computation with a single $2N$-bit adder. Given that the compressor tree's height is roughly $O(\log N)$, the delay of the multiplier can be refined to $O(\log N + \log(2N)) = O(\log N)$. Although Wallace \cite{wallace1964suggestion} and Dadda trees \cite{dadda1965some}—the predominant compressor trees—share a theoretical logarithmic delay, empirical delays vary \cite{townsend2003comparison}, underscoring the impact of the specific tree structure on multiplier performance. Xiao et al. \cite{xiao2021gomil} translated the design of these trees into an integer linear problem, addressed via a combinatorial solver, yet they did not include practical metrics in their model. Zuo et al. \cite{dongsheng2023rl-mul} pioneered the use of reinforcement learning to refine the multiplier design. Their approach, which modifies the Wallace tree structure rather than constructing anew, narrows the state space due to the finite action sequence length. Moreover, the synthesis process remains laborious, presenting challenges in optimizing multipliers exceeding $16$ bits. Furthermore, the technique has not considered the joint optimization of the compressor and prefix trees within the multiplier, which poses a barrier to identifying a globally optimal design.

\subsection{Reinforcement learning}
Reinforcement Learning (RL) has surpassed human performance in many domains, including the ancient game of Go \cite{silver2016mastering}, the complex strategy game StarCraft \cite{vinyals2019grandmaster}, optimizing sorting algorithms \cite{mankowitz2023faster}, and improving matrix multiplication techniques \cite{fawzi2022discovering}. At its heart, RL involves training agents to make a series of decisions to achieve a goal, learning from interactions with their environment by trial and error to maximize a reward over time \cite{liu2022learn,liu2021unsupervised}.
There are two primary categories of RL methods: model-based and model-free. In model-based RL, agents use an explicit model of the environment to inform their decisions. Tools like Monte Carlo Tree Search (MCTS) \cite{swiechowski2023monte}, which simulate various future paths to aid decision-making, are often integrated with these methods. This combination has proven particularly potent for tasks requiring a long sequence of decisions.
Conversely, model-free RL methods, such as DQN \cite{mnih2013playing}, DDPG \cite{lillicrap2015continuous}, and policy gradient approaches \cite{sutton1999policy}, operate without an explicit model of the environment. A prominent example of model-free RL is Proximal Policy Optimization (PPO) \cite{schulman2017proximal, zhuang2023behavior}. This algorithm iteratively refines the agent's policy, optimizing a surrogate objective function to balance the need for stable policy updates with the desire for efficient exploration, leading to high sample efficiency and reduced training times.
Choosing the right RL method is crucial, as different tasks may require different approaches. By aligning the strengths of specific RL techniques with the demands of the task at hand \cite{liu2022dara, liu2024ceil}, agents can navigate complex decision spaces with remarkable effectiveness.

Recent advancements have definitively shown that reinforcement learning is a powerful tool at every phase of hardware design, largely because tasks such as circuit design and testing are fundamentally combinatorial optimization problems. In these tasks, the goal is to navigate a vast solution space to find the most efficient configuration. Notable examples of prior achievements include logic synthesis~\cite{hosny2020drills, zhu2020exploring}, circuit simulation~\cite{niu2023ossp, jin2022accelerating}, chip placement~\cite{mirhoseini2021graph, cheng2021joint, lai2022maskplace, shi2024macro, lai2023chipformer, zhong2024preroutgnn}, chip routing~\cite{chen2023reinforcement, qu2021asynchronous}, clock tree synthesis~\cite{beheshti2021reinforced},  circuit gate sizing~\cite{wang2020gcn, lu2021rl}, and hardware testing~\cite{shi2022deeptpi}, among others.
As such, the widespread success of reinforcement learning in a variety of Electronic Design Automation (EDA) tasks highlights its remarkable capabilities and adaptability as a tool for hardware design optimization.

\end{document}